\documentclass[runningheads]{llncs}

\usepackage{eccv}

\usepackage{eccvabbrv}

\usepackage{graphicx}
\usepackage{booktabs}

\usepackage[accsupp]{axessibility}  

\usepackage[pagebackref,breaklinks,colorlinks,citecolor=eccvblue]{hyperref}

\usepackage{orcidlink}

\usepackage{url}
\usepackage{caption}
\usepackage{subcaption}
\usepackage{multirow}
\usepackage{booktabs}
\usepackage{amssymb}
\usepackage{color,soul}
\usepackage{algorithm}
\usepackage[noend]{algorithmic}
\usepackage{graphicx}
\definecolor{tabhighlight}{HTML}{e5e5e5}
\usepackage{amsmath}
\usepackage[skip=2pt]{caption}
\usepackage{float}
\usepackage{enumerate}

\newcommand{\sbest}[1]{\underline{#1}}
\newtheorem{thm}{Theorem}
\newtheorem{defi}{Definition}

\usepackage{sidecap}
\newcommand{\tablestyle}[2]{\setlength{\tabcolsep}{#1}\renewcommand{\arraystretch}{#2}\centering\footnotesize}

\def\method{SAFT}
\def\FT{FT}
\newcommand{\rotbox}[1]{\rotatebox{55}{#1}}

\usepackage{amsmath,amsfonts,bm}

\def\eqref#1{equation~\ref{#1}}

\def\1{\bm{1}}

\def\ry{{\textnormal{y}}}

\def\rvx{{\mathbf{x}}}

\def\vm{{\bm{m}}}

\def\vt{{\bm{t}}}

\def\vw{{\bm{w}}}
\def\vx{{\bm{x}}}

\def\mI{{\bm{I}}}

\def\mT{{\bm{T}}}

\DeclareMathAlphabet{\mathsfit}{\encodingdefault}{\sfdefault}{m}{sl}
\SetMathAlphabet{\mathsfit}{bold}{\encodingdefault}{\sfdefault}{bx}{n}

\begin{document}

\title{SAFT: Towards Out-of-Distribution Generalization in Fine-Tuning} 

\titlerunning{SAFT: Towards Out-of-Distribution Generalization in Fine-Tuning}

\author{Bac Nguyen\inst{1}\thanks{Correspondence at: \email{bac.nguyencong@sony.com}.}\orcidlink{0000-0001-9193-1908} \and
Stefan Uhlich\inst{2}\orcidlink{0000-0003-3158-4945} \and
Fabien Cardinaux\inst{2}\orcidlink{0000-0003-2921-4873} \and
Lukas Mauch\inst{2}\orcidlink{0000-0001-9212-899X} \and
Marzieh Edraki\inst{3}\orcidlink{0000-0002-1269-1190} \and
Aaron Courville\inst{4,5}\orcidlink{0000-0001-6223-0301}}

\authorrunning{B.~Nguyen et al.}

\institute{Sony AI \and
Sony Europe BV, Stuttgart Laboratory 1 \and
R\&D US Laboratory Sony Corporation of America \and
CIFAR~AI~Chair \and
Mila, Université~de~Montréal}

\maketitle

\begin{abstract}
Handling distribution shifts from training data, known as out-of-distribution (OOD) generalization, poses a significant challenge in the field of machine learning. While a pre-trained vision-language model like CLIP has demonstrated remarkable zero-shot performance, further adaptation of the model to downstream tasks leads to undesirable degradation for OOD data. In this work, we introduce \textbf{S}parse \textbf{A}daptation for \textbf{F}ine-\textbf{T}uning (\method{}), a method that prevents fine-tuning from forgetting the general knowledge in the pre-trained model. \method{} only updates a small subset of important parameters whose gradient magnitude is large, while keeping the other parameters frozen. \method{} is straightforward to implement and conceptually simple. Extensive experiments show that with only 0.1\% of the model parameters, \method{} can significantly improve the performance of CLIP. It consistently outperforms baseline methods across several benchmarks. On the few-shot learning benchmark of ImageNet and its variants, \method{} gives a gain of 5.15\% on average over the conventional fine-tuning method in OOD settings.


\end{abstract}

\section{Introduction}
Visual-language pre-training (VLP) has recently emerged as a powerful method for improving visual representation learning~\cite{radford2021learning,jia2021scaling,pham2023combined}. Simply pre-training on the task of predicting which text goes with which image can substantially improve over image representations learned from scratch~\cite{radford2021learning}. A VLP model consists of an image encoder, a text encoder, and an alignment mechanism for cross-modal interaction. The alignment is often formulated as contrastive learning~\cite{bromley1993signature},  \ie, pulling representations of texts and images which are matched together, while pushing unmatched pairs far away. Besides contrastive loss, learning on large-scale datasets enables VLP models to capture diverse visual concepts~\cite{jia2021scaling}. As a result, they can perform knowledge transfer to many downstream tasks through prompting. A zero-shot classification task can be performed by simply feeding the description of the task-relevant categories to the text encoder and comparing its embeddings with visual embeddings produced by the image encoder. VLP models not only show 
good performance
in-distribution (ID) but also generalize to out-of-distribution (OOD) data.
\begin{figure}[!t]
  \centering
   \includegraphics[width=0.5\linewidth]{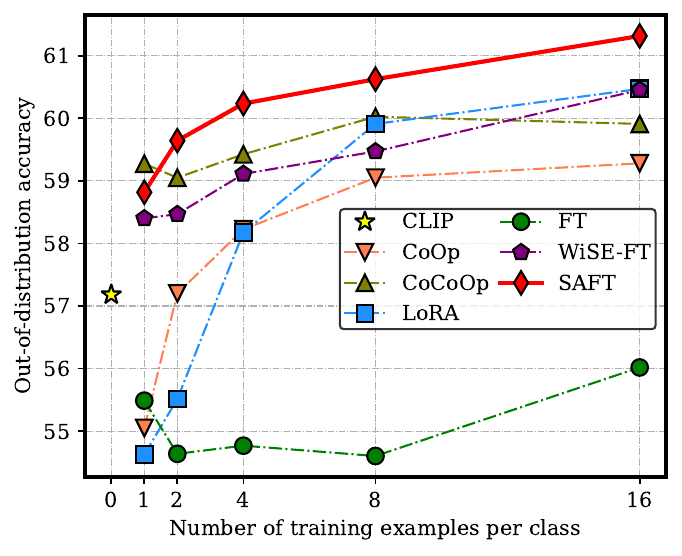}
   \caption{Results for few-shot learning. We report the average accuracy on four distribution-shift variants of ImageNet~\cite{deng2009imagenet}, which are ImageNet-V2~\cite{recht2019imagenet}, ImageNet-Sketch~\cite{wang2019learning}, ImageNet-A~\cite{hendrycks2021natural}, and ImageNet-R~\cite{hendrycks2021many}.}
   \label{fig:nshots-imagenet}
   \vspace{-15pt}
\end{figure}

Nevertheless, to fully leverage VLP capability for a specific downstream task, it is crucial to apply a form of adaptation. Two common adaptation techniques include fine-tuning, which involves optimizing the model parameters, and linear probing, which only adjusts a linear head on top of the frozen pre-trained model. Fine-tuning often leads to higher accuracy compared to linear probing~\cite{kornblith2019better,miller2021accuracy}. However, prior studies have demonstrated that conventional fine-tuning, while enhancing the performance within the same distribution, often leads to decreased robustness against distribution shifts~\cite{kumar2022fine}. This is because an over-parameterized model can easily overfit the specific training data.

Interestingly, despite having a large number of parameters, common pre-trained models tend to have low-dimension reparameterization that is effective for downstream tasks~\cite{aghajanyan2020intrinsic,li2018measuring}. Inspired by this observation, Hu \etal~\cite{hu2021lora} introduced low-rank matrix adaption for fine-tuning large language models. Houlsby \etal~\cite{houlsby2019parameter} proposed lightweight adapter modules with bottleneck structure. We propose a simpler approach compared to finding the low-rank structure. We drastically reduce the number of learnable parameters during fine-tuning, aiming for a minimal impact on VLP models while effectively improving the downstream task performance. Another issue with previous methods is that they are tailored to specific network architectures (\eg transformers~\cite{vaswani2017attention}). Therefore, applying them to various VLP models is not always straightforward .

To address the above issues, this paper introduces an architecture-agnostic framework for parameter-efficient fine-tuning (PEFT).  In particular, we propose to explicitly reduce the number of learnable parameters with \textbf{S}parse \textbf{A}daptation for \textbf{F}ine-\textbf{T}uning (\textbf{\method{}}). Instead of updating all model parameters, \method{} identifies a subset of learnable parameters that are effective for a specific downstream task. As an illustrative example, \cref{fig:nshots-imagenet} shows the comparison of \method{} against other baseline methods in terms of OOD accuracy. The training dataset is ImageNet~\cite{russakovsky2015imagenet} and the validation is conducted on distribution-shift variants of ImageNet (see~\cref{sub:domain_shift} for more details). Under different few-shot learning settings, our method consistently and significantly demonstrates superior generalization capabilities.  In summary, this paper makes the following contributions.
\begin{enumerate}[(i)]
    \item We introduce \method{}, a simple and effective method for adapting VLP models, that can achieve better OOD performance than conventional fine-tuning (\cref{sub:saft}). Our key idea is to only adapt a small subset of learnable parameters that are important for downstream fine-tuning. As a result, the fine-tuned model can preserve the OOD generalization capability of the pre-trained model.
    \item We conduct experiments of fine-tuning CLIP~\cite{radford2021learning} on common benchmarks, including distribution shifts (\cref{sub:domain_shift}), generalization to new classes (\cref{sub:generalization_new_classes}), and cross-dataset transfer evaluation (\cref{sub:generalization_across_dataset}). We provide several ablation studies to give deeper insight into the proposed method (\cref{sub:ablation}).
    \item We further extend our evaluation to language models applied to natural language processing (NLP) tasks (\cref{sub:nlp}). In particular, \method{} outperforms Low-rank Adapter~\cite{hu2021lora} (LoRA) in four out of five tasks in terms of OOD performance. The results show that \method{} is a model-agnostic method and can be applied to broader contexts beyond VLP models.
\end{enumerate}

\section{Sparse Adaptation for Fine-Tuning}
\label{sec:method}
In this section, we introduce \method{}. For the sake of simplicity, we formalize it based on the CLIP model. Please note that \method ~can be easily extended to other modalities and tasks as we  show in~\cref{sub:nlp}.

\begin{figure}[t]
  \centering
   \includegraphics[width=\linewidth]{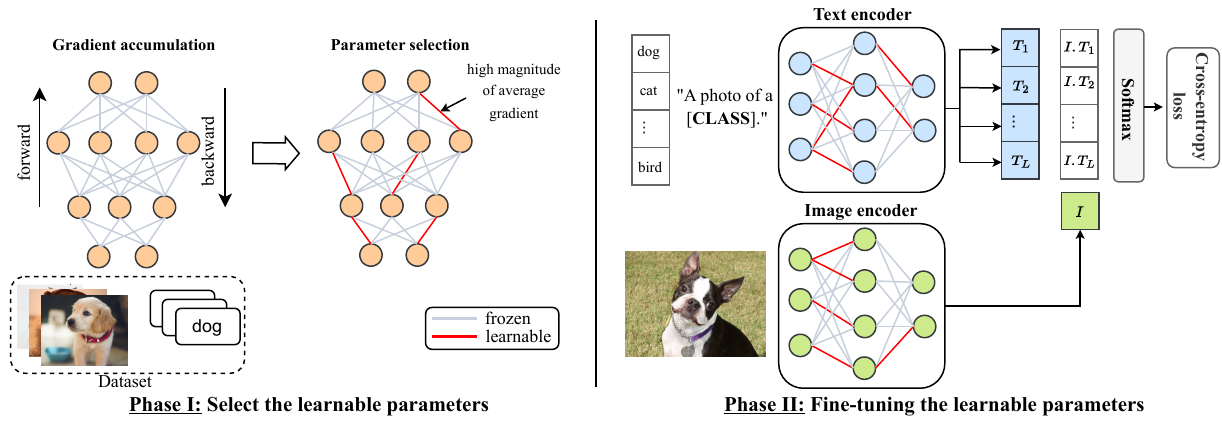}
   \caption{An overview of \textbf{S}parse \textbf{A}daptation for \textbf{F}ine-\textbf{T}uning (\method{}). Our method consists of two phases: (I) We use the downstream dataset to select learnable parameters; (II) We fine-tune the model on the downstream dataset.}
   \label{fig:method}
   \vspace{-10pt}
\end{figure}

\subsection{Preliminaries}
\textbf{Problem setup.}
Consider a supervised learning setting for a downstream task, where training data $\{(\vx_i, y_i)\}_{i=1}^N$  are samples from a distribution $\mathbb{P}_{\text{ID}}(\rvx,\ry)$.  Here, $\vx_i$ denotes a raw image, and $y_i$ denotes the corresponding class label. Our objective is to fine-tune a pre-trained CLIP~\cite{radford2021learning} model for this downstream task in such a way that it can achieve robust generalization on related but OOD data. We evaluate the robustness of the fine-tuned model in two different scenarios: \textit{in-distribution} (ID)  and \textit{out-of-distibution} (OOD). In ID settings, test examples are independently and identically distributed samples from the same training distribution $\mathbb{P}_{\text{ID}}(\rvx, \ry)$. In OOD settings, test examples are drawn from a distribution $\mathbb{P}_{\text{OOD}}(\rvx, \ry)$, which is different from $\mathbb{P}_{\text{ID}}(\rvx, \ry)$. The latter scenario is more challenging since the model must be capable of generalizing beyond the training distribution.

\textbf{CLIP Fine-Tuning.}
The visual-language CLIP model consists of an image encoder $g_{\theta_{I}}$ and a text encoder $h_{\theta_{T}}$. Let $\mI_{\vx} = g_{\theta_{I}}(\vx) / \|g_{\theta_{I}}(\vx)\|_2$ and $\mT_{\vt} = h_{\theta_{T}}(\vt)/\|h_{\theta_{T}}(\vt)\|_2$ denote the normalized output embedding of an image $\vx$  and text $\vt$, respectively. Let $\theta=[\theta_{I},\theta_{T}]^\top \in \mathbb{R}^D$ be the model parameters, including both image and text encoders. CLIP formulates the learning objective as a contrastive loss, which pulls together matching text and image representations while pushing apart unmatched representations. By pre-training at a large scale, CLIP can learn diverse visual concepts, making it transferable to many downstream tasks through prompting~\cite{radford2021learning,zhou2022conditional}. To perform zero-shot classification, we first transform the class label into a descriptive text prompt, \eg, ``\texttt{a photo of a [CLASS].}'', where the ``\texttt{[CLASS]}'' token is replaced by the actual class name. Let ${\vt_y}$ denote the prompt corresponding to class label $y$. The cosine similarity between an image $\vx$ and a class label $y$ is computed as $f_\theta(\vx, y) = \mI_{\vx}^\top \mT_{\vt_y}$. The prediction probability of $\vx$ belonging to class $y$ is computed as
\begin{align}
    \mathbb{P}(\ry = y| \vx; \theta) = \frac{\exp( f_\theta(\vx,y) / \tau)}{\sum_{c=1}^{L} \exp( f_\theta(\vx, c)/ \tau)}\,,
\end{align}
where $L$ denotes the number of class labels in the test set and $\tau > 0$ denotes the temperature. Although the contrastive loss is used to pre-train CLIP, we simply consider the cross-entropy (CE) loss $\mathcal{L}_{\text{CE}}(\vx_i, y_i; \theta) = -\log \mathbb{P}(y_i| \vx_i; \theta)$ to fine-tune CLIP as commonly done in previous studies~\cite{zhou2022conditional,zhou2022learning,shu2023clipood,wortsman2022robust}.

\subsection{Proposed Method} \label{sub:saft}
\textbf{Motivation.}
\method{} is motivated by the recent success of PEFT for large pre-trained models~\cite{zhang2021tip,hu2021lora,aghajanyan2020intrinsic,guo2021parameter,fu2023effectiveness,sung2021training}. Instead of updating all model parameters, we only fine-tune a critical subset of the parameters, specifically those pertinent to the downstream task. Our work draws parallels to the concept of pruning in neural network compression where only a crucial subset of parameters is retained~\cite{cheng2023survey}. The parameters identified by our method can be seen as a subnetwork with a better OOD inductive bias, as conjectured by the Lottery Ticket Hypothesis~\cite{frankle2018lottery,zhang2021can}.  Existing PEFT methods, though effective, face challenges in selecting the appropriate parameters. They often rely on pre-defined rules (e.g., only updating the biases~\cite{zaken2022bitfit} or low-rank adaptation~\cite{hu2021lora}) without leveraging task-specific data. Although a few attempts~\cite{guo2021parameter,fu2023effectiveness} have been made to address this issue, these methods either come with high computational costs or have limitations. The key differences in our work compared to previous PEFT methods are as follows.
(1) We introduce a straightforward approach for parameter selection, marking a step forward from existing methods that are computationally intensive or limited in scope. (2) While previous methods focus on conventional ID fine-tuning, we aim to improve OOD performance. \method{} is based on recent theoretical insights~\cite{arora2018stronger, aghajanyan2020intrinsic} and empirical studies indicating that reducing the number of learnable parameters can enhance OOD generalization. More specifically, we derive the generalization bound for \method{}  based on a generalization bound framework via compression (see~\cref{sub:bound}). This generalization bound depends on the number of learnable parameters, which is much smaller for \method{} than the number of parameters of the pre-trained models. 

\textbf{Method.} \method{} consists of two phases (see \cref{fig:method}). The first phase involves the selection of learnable parameters. The second phase adapts the selected parameters for downstream tasks. A pseudocode of \method{} is given in Algorithm~\ref{alg:spa}.

In the first phase, we identify the most important parameters for the downstream task. While our proposed method can be applied to any training objective, we focus on classification as an illustrative use case. In particular, we compute the gradient of the CE loss function\footnote{Please note that SAFT does not require specifically the CE loss. It can easily be extended to other loss functions, depending on the downstream tasks.} with respect to the parameters, \ie, $\nabla_{\theta} \mathcal{L}_{\text{CE}}(\vx_i, y_i; \theta)$. These gradients are then averaged over the data,
\begin{equation}
    \vw = \frac{1}{N} \sum\nolimits_{i=1}^N \nabla_{\theta} \mathcal{L}_{\text{CE}}(\vx_i, y_i; \theta)\,. \label{eq:importance}
\end{equation}
Essentially, $\vw$ quantifies how much the loss function changes in response to a small change in each parameter. By selecting only the parameters $\theta_{i}$ with high values $|w_i|$, we maximize the effect on the loss function while reducing drastically the number of learnable parameters. Parameters with the largest gradient magnitudes yield fast convergence on the downstream task. Moreover, restricting the number of learnable parameters ensures that the fine-tuned model is close to the pre-trained model. A similar technique to compute the parameter importance was proposed in Elastic Weight Consolidation~\cite{kirkpatrick2017overcoming}. However, instead of computing the mean of gradient magnitudes (the diagonal of the Fisher information matrix), we average gradients over the given examples and then compute the magnitudes. \method{} has the advantage of mitigating the influence of outliers.

\method{} can be implemented by a masking strategy. Let $\alpha\in [0,1]$ denote the sparsity level. When $\alpha = 0$, it corresponds to the pre-trained model without learnable parameters. When $\alpha = 1$, it corresponds to the conventional fine-tuning approach, where all parameters are learnable. By varying the sparsity level, we can control the number of learnable parameters. Let $\vm\in \{0,1\}^D$ denote the mask for learnable parameters, which can be computed as
\begin{equation}
m_k = \left\{\begin{array}{lr}
    1, & \text{if $|w_k| \ge {\text{sorted}}(|w|_1, \dots, |w|_D)[d]$}\\
    0, & \text{otherwise}\\
    \end{array}\right. \,,\label{eq:mask}
\end{equation}
where $\text{sorted}(.)$ denotes the sorting function of an array in descending order and $d=\lfloor \alpha D \rfloor$ denotes the number of learnable parameters. It is important to note that the mask is not updated during fine-tuning.

In the second phase, we fine-tune the masked parameters. The CE loss function is minimized using stochastic gradient descent. In each iteration, after an update, we reset unmasked parameters to the values of the pre-trained model. By updating only the masked parameters, we expect the model to focus only on relevant features that are useful to solve the downstream task. Since the other parameters of the network are frozen, this approach can prevent fine-tuning from forgetting the general knowledge retained from the pre-trained model.

\begin{figure}[t]
    \centering
    \includegraphics[width=0.49\linewidth]{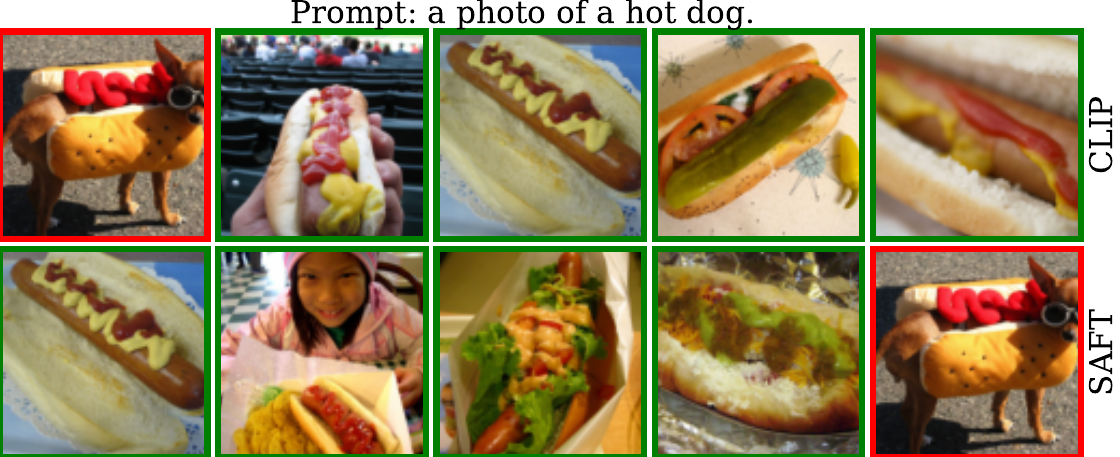}
    \includegraphics[width=0.49\linewidth]{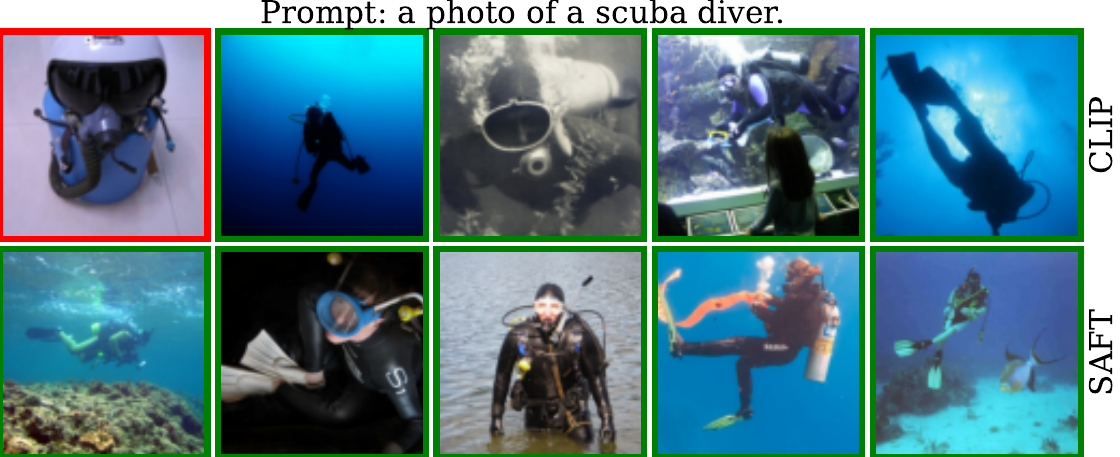}
    \caption{Top-5 retrieved images for a given prompt. Images are arranged from left to right in descending order of similarity to the given prompt. A green box indicates a correct match between image and text, while a red box indicates an incorrect match.}
    \label{fig:image_retrieval}
    \vspace{-10pt}
\end{figure}

As an illustrative example, we show the top-5 image retrieval results on the test set of ImageNet in \cref{fig:image_retrieval}. Training data consists of 16 shots for each class from ImageNet. \method{} successfully retrieves the most relevant images for a given prompt, whereas CLIP may obtain incorrect matches.

\setlength{\textfloatsep}{10pt}
\begin{algorithm}[tb]
\caption{\textbf{S}parse \textbf{A}daptation for \textbf{F}ine-\textbf{T}uning (\method{})}
\label{alg:spa}
\begin{algorithmic}
\STATE \textbf{Input:} Data for fine-tuning $\{(\vx_i, y_i)\}_{i=1}^N$; a pre-trained model $f_\theta(.,.)$;
\STATE \qquad \quad sparsity level $0 \le \alpha \le 1$; number of iterations $T$
\STATE \textbf{\underline{Phase 1:}} Select a subset of learnable parameters
\STATE \quad $\vw \leftarrow \mathbf{0}$
\STATE \quad \textbf{For} $i\leftarrow 1$ to $N$
\STATE \quad \quad $\vw \leftarrow \vw + (1/N)\nabla_{\theta} \mathcal{L}_{\text{CE}}(\vx_i, y_i; \theta)$
\STATE \quad \textbf{End For}
\STATE \quad Compute $\vm$ for sparsity level $\alpha$ using \cref{eq:mask}
\STATE \textbf{\underline{Phase 2:}} Fine-tune the learnable parameters
\STATE \quad $\tilde{\theta}\leftarrow \theta$
\STATE \quad \textbf{For} $t\leftarrow 1$ to $T$
\STATE \quad \quad Update $\tilde{\theta}$ using gradient for mini-batch $\mathcal{B}_t$
\STATE \quad \quad \quad \quad $(1/|\mathcal{B}_t|)\sum_{(\vx_i, y_i)\in \mathcal{B}_t}\nabla_{\theta} \left . \mathcal{L}_{\text{CE}}(\vx_i, y_i; \theta) \right |_{\theta = \tilde{\theta}}$
\STATE \quad \quad Reset unmasked parameters $\tilde{\theta} \leftarrow \vm\odot\tilde{\theta} + (1 - \vm)\odot\theta$
\STATE \quad \textbf{End For}
\STATE {\textbf{Return:} Fine-tuned model $f_{\tilde{\theta}}(.,.)$}
\end{algorithmic}
\end{algorithm}
\vspace{-10pt}

\section{Experiments}
\begin{table}[t!]
\centering
\resizebox{\linewidth}{!}{%
\begin{tabular}{lrcccccc}
\toprule
 \multirow{3}{*}{Method} & \multirow{3}{*}{\shortstack{\# fine-tuned\\ parameters}} & ID & \multicolumn{5}{c}{OOD}\\
 \cmidrule(lr){3-3} \cmidrule(lr){4-8}
 & & ImageNet & ImageNet-V2 & ImageNet-S & ImageNet-A & ImageNet-R & \textit{Average} \\
\midrule
CLIP~\cite{radford2021learning}  & 0 & 66.73 & 60.83 & 46.15 & 47.77 & 73.96 & 57.18 \\
CoOp~\cite{zhou2022learning}   & 2048 & 71.51 & 64.20 & 47.99 & 49.71 & 75.21 & 59.28 \\
CoCoOp~\cite{zhou2022conditional} & 35360 & 71.02 & 64.07 & 48.75 & 50.63 & 76.18 & 59.91 \\
MaPLe~\cite{khattak2023maple} & 3.55 M & 70.72 & 64.07 & 49.15 & \sbest{50.90} & 76.98 & 60.28 \\
CLIPood~\cite{shu2023clipood} & 86.19 M & 71.60 & 64.90 & 49.30 & 50.40 & \sbest{77.20} & 60.40 \\
LoRA~\cite{hu2021lora} & 0.12 M & 72.68 & 65.57 & 48.61 & 49.39 & 76.29 & 60.47 \\
WiSE-FT~\cite{wortsman2022robust} & 149.62 M & \textbf{73.91} & \textbf{66.69} & \sbest{49.67} & 49.00 & 77.11 & \sbest{60.62} \\
LP-FT~\cite{kumar2022fine} & 87.22 M & 71.65 & 62.69 & 43.35 & 40.49 & 69.77 & 54.08 \\
FLYP~\cite{goyal2023finetune} & 149.62 M & 73.26 & 66.35 & 49.21 & 49.40 & 77.13 & 60.52 \\
\FT{} & 149.62 M & 71.55 & 63.59 & 45.25 & 42.32 & 73.52 & 56.17 \\
\midrule
\method{} & 0.15 M & \sbest{72.71} & \sbest{65.84} & \textbf{49.69} & \textbf{51.60} & \textbf{78.13} & \textbf{61.32} \\
\bottomrule
\end{tabular}}
\caption{Classification accuracy (\%) for distribution shifts on ImageNet. The \textbf{best} and \sbest{second best} results on each dataset are marked.} \label{table:imagenet_distribution_shift}
\vspace{-10pt}
\end{table}

We conduct extensive experiments to validate the robustness of \method{} across multiple benchmarks, including distribution shifts, generalization from base to new classes, and cross-dataset transfer. For VLP model, all experiments are based on the open-source implementation of CLIP\footnote{Available at \url{https://github.com/openai/CLIP}}. We follow the same evaluation protocol as suggested by Zhou  \etal~\cite{zhou2022conditional}. A few-shot training strategy is employed by randomly selecting 16 shots for each class. Unless otherwise specified, image encoder ViT-B/16~\cite{dosovitskiy2020image} is used as a default choice.

\textbf{Implementation details.} We employ the pre-trained model CLIP as the backbone network. Unless specified differently, the sparsity level $\alpha$ is set to 0.001 for all datasets. \method{} is trained using the AdamW~\cite{loshchilov2017decoupled} optimizer with a weight decay of 0.1. As a default setting, we use a learning rate of $5\times 10^{-6}$ with the cosine learning rate strategy. For \method{}, we fine-tune both image and text encoders. Experiments are conducted on a single NVIDIA RTX 6000 ADA GPU with 48GB of memory. Both training and inference are conducted with mixed precision of bfloat16 (brain floating‐point).

\textbf{Competing methods.} \method{} is compared against several state-of-the-art methods for adapting  VLP models. As a baseline, we consider the conventional fine-tuning method (FT), which employs gradient descent to update all the model parameters using cross-entropy loss. More recent fine-tuning methods include WiSE-FT~\cite{wortsman2022robust} and CLIPood~\cite{shu2023clipood}, which rely on parameter ensembling. In WiSE-FT, a simple linear interpolation is applied to the pre-trained and fine-tuned parameters, while CLIPood maintains a temporal ensemble weighted by the Beta distribution. FLYP~\cite{goyal2023finetune} is a simple approach that mimics the contrastive pre-training for fine-tuning.  LP-FT~\cite{kumar2022fine} performs a two-stage fine-tuning process where linear probing is first performed and then full fine-tuning.  Additionally, we report the performance of LoRA~\cite{hu2021lora}, a widely used technique for fine-tuning large language models.  We also compare \method{} against various prompt-tuning methods, including CoOp~\cite{zhou2022learning}, CoCoOp~\cite{zhou2022conditional}, and MaPLE~\cite{khattak2023maple}. For more details about these competing methods, please refer to~\cref{appendix:baseline}.

\subsection{Generalization to Distribution Shifts} \label{sub:domain_shift}

\textbf{Benchmark.} We consider the following benchmark, where ImageNet~\cite{deng2009imagenet} is used for fine-tuning. The ID evaluation is employed on the test set of ImageNet and the OOD evaluation is employed on ImageNet-V2~\cite{recht2019imagenet}, ImageNet-Sketch~\cite{wang2019learning}, ImageNet-A~\cite{hendrycks2021natural}, and ImageNet-R~\cite{hendrycks2021many}. These datasets are different variants of ImageNet, in which the class labels align with those of ImageNet and there are no training examples available. Please refer to~\cref{appendix:dataset} for more details.

\textbf{Results.} \cref{table:imagenet_distribution_shift} presents both ID and OOD generalization results. \method{} demonstrates comparable performance to state-of-the-art methods on ID data, while achieving the best OOD performance on average, highlighting its strong generalization capabilities across various distribution shifts. Notably, all methods yield improvements over CLIP in terms of ID performance.  SAFT outperforms traditional fine-tuning (FT) by a significant margin of 5.15\% on OOD data, indicating that sparse adaptation effectively mitigates overfitting during the fine-tuning process. Even when compared to the competitive ensemble methods WiSE-FT and CLIPood, our approach continues to achieve superior results. Although WiSE-FT can be straightforwardly applied to \method{}, our preliminary studies showed that we did not observe any improvements. Moreover, \cref{fig:nshots-imagenet} shows the results under different numbers of shots. We use the same splits for training and evaluation as suggested by Zhou \etal~\cite{zhou2022learning}. \method{} achieves the best OOD performance in most settings, even when dealing with very few shots. 

\subsection{Generalization from Base to New Classes} \label{sub:generalization_new_classes}

\begin{figure}[tb]
    \centering
    \begin{subfigure}[b]{0.46\textwidth}
         \centering
    \includegraphics[width=\textwidth]{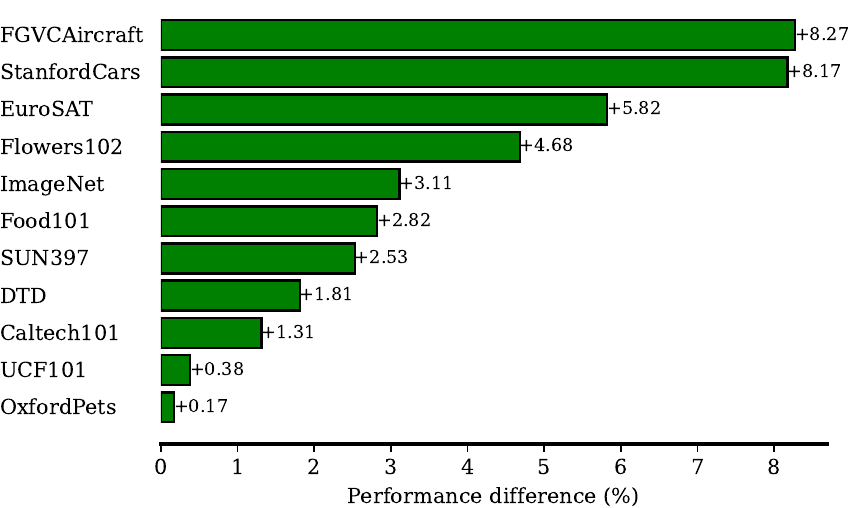}
    \caption{New classes}
    \end{subfigure}
    \centering
    \begin{subfigure}[b]{0.46\textwidth}
   \includegraphics[width=\textwidth]{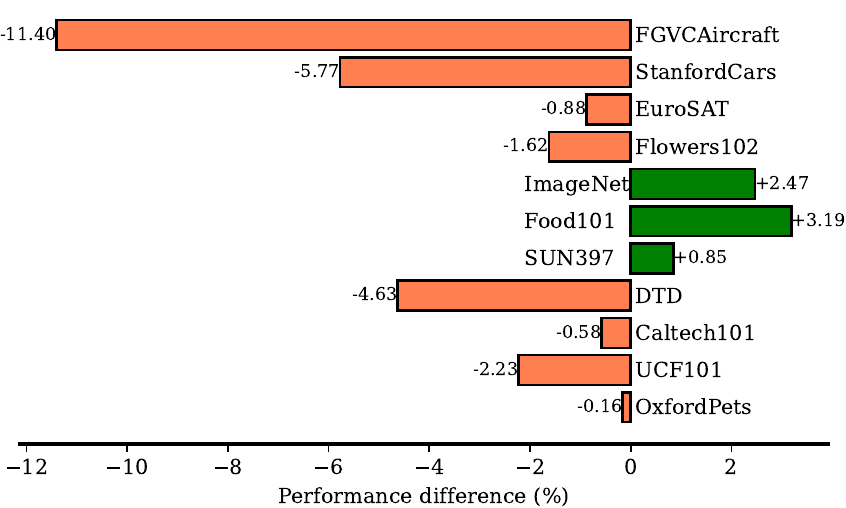}
   \caption{Base classes}
   \end{subfigure}%
   \caption{Performance difference in base-to-new generalization settings. We report the difference between \method{} and FT: (a) in new classes and (b) in base classes.}
   \label{fig:absolute}
   \vspace{-5pt}
\end{figure}


\textbf{Benchmark.} Following Zhou  \etal~\cite{zhou2022conditional}, we use a benchmark of 11 datasets (see~\cref{appendix:dataset} for more details), covering a wide range of recognition tasks. The benchmark contains ImageNet~\cite{deng2009imagenet} and Caltech101~\cite{fei2004learning} for generic object classification; OxfordPets~\cite{parkhi2012cats},  StanfordCars~\cite{krause20133d}, Flowers102~\cite{nilsback2008automated}, Food101~\cite{bossard2014food}, and FGVCAircraft~\cite{maji2013fine} for fine-grained classification; SUN397~\cite{xiao2010sun} for scene recognition; UCF101~\cite{soomro2012ucf101} for action recognition; DTD~\cite{cimpoi2014describing} for texture classification; and EuroSAT~\cite{helber2019eurosat} for satellite imagery recognition. The classes are equally divided into two groups, with one group serving as base classes and the other as new classes. As noted by Zhou \etal~\cite{zhou2022conditional}, the split does not ensure that the two groups of classes have equal difficulty levels. All competing methods are trained only in base classes and then subsequently evaluated in both base and new classes separately.  It is important to note that the test data may include new classes that were not seen in the training data.

\begin{table}[!t]
\centering
\setlength{\tabcolsep}{10pt}
\resizebox{0.7\linewidth}{!}{
\begin{tabular}{lcccccc}
\toprule
\multirow{3}{*}{Method}   & \multicolumn{3}{c}{\textit{Average}} & \multicolumn{3}{c}{ImageNet} \\
\cmidrule(lr){2-4} \cmidrule(lr){5-7}
 & Base  & New   & H   & Base  & New   & H     \\
\midrule
CLIP~\cite{radford2021learning}    & 69.34 & 74.22 & 71.70  & 72.43 & 68.14 & 70.22 \\
CoOp~\cite{zhou2022learning}   & 82.69 & 63.22 & 71.66  & 76.47 & 67.88 & 71.92\\
CoCoOp~\cite{zhou2022conditional} & 80.47 & 71.69 & 75.83  & 75.98 & 70.43 & 73.10 \\
MaPLe~\cite{khattak2023maple} & 82.28 & \textbf{75.14} & 78.55  & 76.66 & \sbest{70.54} & 73.47 \\
CLIPood~\cite{shu2023clipood}  & 83.90 & 74.50 & \sbest{78.90}  & \sbest{77.50} & 70.30 & \sbest{73.70}\\
\FT{} & \textbf{85.86}  & 71.23 & 77.86 &  75.27 & 68.29 &  71.61 \\
\midrule
\method{} & \sbest{83.97} &  \sbest{74.78} &  \textbf{79.11} & \textbf{77.74}  & \textbf{71.40} & \textbf{74.44} \\
\bottomrule                          
\end{tabular}}
\caption{Classification accuracy (\%) from base to new classes over 11 datasets. The \textbf{best} and \sbest{second best} results are marked.}
\label{table:openclass}
\vspace{-10pt}
\end{table}

\textbf{Results.}  We present the classification accuracy for both base classes (Base) and new classes (New), along with their harmonic mean~\cite{xian2017zero} ($\text{H} = 2 \times \text{Base} \times \text{New}/(\text{Base} + \text{New})$), to emphasize the trade-off between downstream adaptation and new-class generalization. We report the results comparing \method{} with CLIP, CoOp, CoCoCop, MaPLE, CLIPood, and conventional FT. \cref{table:openclass} shows the generalization results from base to new classes. For reference, comprehensive results on each dataset are provided in~\cref{appendix:base_to_new}. As we can see, SAFT obtains significant improvement over CLIP in base class settings, indicating its capacity to enhance the model performance in downstream tasks. More importantly, \method{} preserves its OOD generalization ability to new classes. In particular, our method improves the accuracy in base classes from 69.34\% to 83.97\% and in new classes from 74.22\% to 74.78\%, giving the best harmonic mean accuracy of 79.11\%. On ImageNet, Food101, and SUN397, \method{} outperforms FT in both base and new classes. This could be explained by the overfitting of FT for base classes. As expected, the conventional FT achieves the best results in base classes, but it does not generalize well on unseen classes. FT gives worse results than CLIP in new classes. Importantly, our method performs the best for both base and new classes on ImageNet. Furthermore, \cref{fig:absolute} shows the performance differences of \method{} over FT on each dataset for both base and new classes. Our method consistently improves the results in new class settings across all datasets. Since FT  has more degrees of freedom to fit the training data, it can obtain better performance than \method{} in base classes. Overall, \method{} still brings positive improvement over FT.

\subsection{Cross-Dataset Transfer} \label{sub:generalization_across_dataset} 

\begin{table}[t]
\small
\setlength{\tabcolsep}{4pt}
\renewcommand{\arraystretch}{0.9}
\resizebox{\linewidth}{!}{%
\begin{tabular}{lcccccccccccc}
\toprule
 \multirow{3}*{Method} & Source & \multicolumn{11}{c}{Target} \\ \cmidrule(lr){2-2} \cmidrule(lr){3-13}
& \rotbox{ImageNet} & \rotbox{Caltech101} & \rotbox{OxfordPets} & \rotbox{StanfordCars} & \rotbox{Flowers102} & \rotbox{Food101} & \rotbox{FGVCAircraft} & \rotbox{SUN397} & \rotbox{DTD} & \rotbox{EuroSAT} & \rotbox{UCF101} & \rotbox{\emph{Average}} \\
\midrule
CLIP~\cite{radford2021learning}  & 66.73 & 93.59 & 89.15 & \sbest{65.41} & 70.36 & 85.58 & 24.33 & 63.05 & 43.32 & 46.51 & 67.14 & 64.85 \\
CoOp~\cite{zhou2022learning}  & 71.51 & 93.70 & 89.14 & 64.51 & 68.71 & 85.30 & 18.47 & 64.15 & 41.92 & 46.39 & 66.55 & 63.88 \\
CoCoOp~\cite{zhou2022conditional} & 71.02 & \textbf{94.43} & 90.14 & 65.32 & \sbest{71.88} & 86.06 & 22.94 & \sbest{67.36} & 45.73 & {45.37} & 68.21 & 65.74 \\
MaPLe~\cite{khattak2023maple} & 70.72 & 93.53 & \sbest{90.49} & \textbf{65.57} & \textbf{72.23} & \sbest{86.20} & \sbest{24.74} & 67.01 & \textbf{46.49} & \sbest{48.06} & \sbest{68.69} & \sbest{66.30}\\
\FT{} & \sbest{71.55} & 93.10 & 86.32 & 57.80 & 67.68 & 81.01 & 21.06 & 64.87 & 43.14 & 34.81 & 67.30 & 61.71  \\
\midrule
\method{} & \textbf{72.71} &  \sbest{94.16} & \textbf{90.87} & 65.34 & 70.77 & \textbf{86.27} & \textbf{25.17} & \textbf{68.15} & \sbest{46.16} & \textbf{49.58} & \textbf{70.21} & \textbf{66.67}\\
\bottomrule
\end{tabular}}
\caption{Classification accuracy (\%) in cross-dataset transfer setting. The \textbf{best} and \sbest{second best} results are marked.}
\label{tab:xd}
\vspace{-10pt}
\end{table}


\textbf{Benchmark.}  We evaluate \method{} cross-dataset generalization ability by fine-tuning it on ImageNet and subsequently applying this learning directly to the other 10 datasets, including Caltech101~\cite{fei2004learning}, OxfordPets~\cite{parkhi2012cats},  StanfordCars~\cite{krause20133d}, Flowers102~\cite{nilsback2008automated}, Food101~\cite{bossard2014food}, FGVCAircraft~\cite{maji2013fine}, SUN397~\cite{xiao2010sun}, UCF101~\cite{soomro2012ucf101}, DTD~\cite{cimpoi2014describing}, and EuroSAT~\cite{helber2019eurosat}. These datasets have been used for benchmarking in \cref{sub:generalization_new_classes}. However, in this context, we consider all classes for evaluation. 

\textbf{Results.} \cref{tab:xd} shows the performance comparison between \method{}, CoOp, CoCoOp, MaPLe, and FT. On the source dataset of ImageNet, \method{} obtains the best performance, followed by FT and CoOp.  Furthermore, our method obtains considerably stronger generalization by outperforming other state-of-the-art methods on 6 out of 10 datasets. On average, SAFT shows competitive performance, resulting in the highest average accuracy of 66.67\%. It shows a gain of 1.82\% over CLIP on unseen datasets. On SUN397, \method{} gives a gain of more than 5\%.

\subsection{Ablation Studies} \label{sub:ablation}

\textbf{Parameter selection strategy.} Our approach relies on gradient magnitudes for the selection of learnable parameters. To highlight the significance of this selection strategy, we compare SAFT against two other strategies. The first straightforward strategy involves a random selection of learnable parameters from the pre-trained model (\textit{Random}). The second strategy involves the selection of parameters with the smallest magnitudes (\textit{WM}), which is widely used in network pruning~\cite{han2015learning}. These parameters have negligible effects between layers, which help the fine-tuned model more robust to OOD. To make a fair comparison, we make sure that all these methods have approximately an equal number of learnable parameters. Specifically, we use 0.1\% of parameters as learnable parameters to be comparable to \method{}. \cref{tab:selection_strategy} shows the results on the ImageNet benchmark. As shown in the table, SAFT obtains superior performance than the other strategies, demonstrating the effectiveness of using gradient magnitudes for parameter selection. The results obtained using random selection further indicate that sparsity is not the sole factor contributing to the success of our approach. It is crucial to select the right parameters.

\begin{table}[tb]
\centering
\resizebox{0.7\linewidth}{!}{
\begin{tabular}{lccccc}
\toprule
 \multirow{3}*{Method} & ID & \multicolumn{4}{c}{OOD}\\
 \cmidrule(lr){2-2} \cmidrule(lr){3-6}
 & ImageNet & ImageNet-V2 & ImageNet-S & ImageNet-A & ImageNet-R \\
\midrule
\method{} & \textbf{72.71} & \textbf{65.84} & \textbf{49.69} & \textbf{51.60} & \textbf{78.13}  \\
\midrule
Random & 68.77 & 62.19 & 47.72 & 48.37 & 75.42  \\
WM & 67.56 & 61.13  & 46.94 & 47.45 & 74.36 \\
\bottomrule
\end{tabular}}
\caption{Ablation studies on the ImageNet benchmark. The best results are in \textbf{bold}.}
\label{tab:selection_strategy}
\vspace{-5pt}
\end{table}

\begin{figure}[!t]
\begin{minipage}[c]{0.5\linewidth}
   \includegraphics[width=\linewidth]{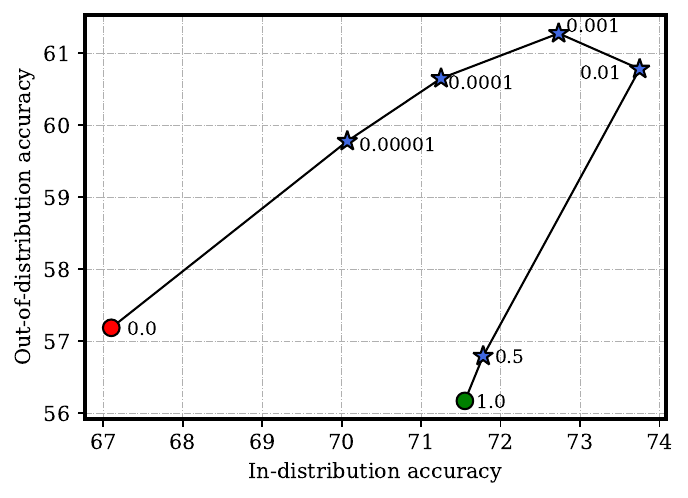}
   \caption{ID vs OOD performance with different sparsity levels.}
   \label{fig:change_rate}
\end{minipage}
\hfill
\begin{minipage}[c]{0.45\linewidth}
    \begin{subfigure}[b]{\linewidth}
       \centering
       \includegraphics[width=0.7\linewidth]{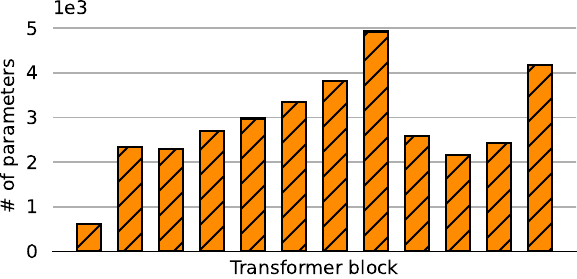}
       \caption{Image encoder}
   \end{subfigure}
   \hfill
   \begin{subfigure}[b]{\linewidth}
       \centering
       \includegraphics[width=0.7\linewidth]{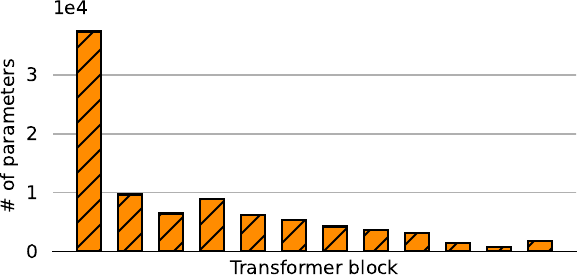}
       \caption{Text encoder}
   \end{subfigure}
   \captionof{figure}{Number of trainable parameters selected by \method{} per block for (a)~image encoder and (b)~text encoder.}
   \label{fig:selected_params}
\end{minipage}
\end{figure}

\textbf{Sparsity level.}  To see the effects of sparsity level, we evaluate the generalization of \method{} for different values. \cref{fig:change_rate} depicts the  ID against OOD performance of \method{}  when changing the sparsity levels. As expected, increasing the sparsity level enhances the ID performance, but it also decreases the OOD performance.  In~\cref{appendix:nlp}, we further illustrate the influence of $\alpha$ on fine-tuning language models on NLP tasks. As shown in the results, given that $\alpha$ is relatively small, while the optimal sparsity levels can slightly vary across different tasks and models, setting $\alpha = 0.1\%$ gives good overall performance.



\textbf{Image encoder.} We evaluate the effectiveness of \method{} for different encoders. In particular, the architecture for the image encoder in CLIP can be based on a Convolutional Neural Network (CNN) such as ResNet-50~\cite{he2016deep} or a Vision Transformer (ViT) as proposed by Dosovitskiy \etal~\cite{dosovitskiy2020image}. We report the performance of \method{} against CLIP, Linear Probing (LP), CoOP, and FT  considering different vision backbones. The results of distribution shifts on the ImageNet benchmark are summarized in \cref{tab:different_backbon}. As indicated in the table, \method{} consistently enhances the OOD performance and achieves the top result among the competing methods. This suggests that \method{} is architecture agnostic.

\begin{table}[!t]
\resizebox{\linewidth}{!}{%
\renewcommand{\arraystretch}{0.8}
		\label{table:student}
		\centering
		\begin{tabular}{llcccccc}
\toprule
 \multirow{3}{*}{Architecture} & \multirow{3}{*}{Method} & ID & \multicolumn{5}{c}{OOD}\\
 \cmidrule(lr){3-3} \cmidrule(lr){4-8}
 & & ImageNet & ImageNet-V2 & ImageNet-S & ImageNet-A & ImageNet-R & \textit{Average} \\
\midrule
 \multirow{2}{*}{ResNet-50} & CLIP~\cite{radford2021learning} & 58.18 & 51.34 & 33.32 & 21.65 & 56.00 & 40.58 \\
 & LP & 55.87 & 45.97 & 19.07 & 12.74 & 34.86 & 28.16 \\
 & CoOp~\cite{zhou2022learning} & \textbf{63.33} & \textbf{55.40} & \sbest{34.67} & \textbf{23.06} & \sbest{56.60} & \sbest{42.43} \\
 & \FT{} & 60.95 & 52.25 & 28.21 & 15.73 & 50.06 & 36.56\\
 \cmidrule(lr){2-8}
 & \method{} & \sbest{62.80} & \sbest{55.22} & \textbf{34.93} & \sbest{23.00} & \textbf{59.86} & \textbf{43.25} \\
 \midrule
 \multirow{2}{*}{ResNet-101}  & CLIP~\cite{radford2021learning} & 61.62 & 54.81 & 38.71 & 28.05 & 64.38 & 46.49 \\
  & LP & 59.75 & 50.05 & 26.80 & 19.44 & 47.19 & 35.87 \\
 & CoOp~\cite{zhou2022learning} & \textbf{65.98} & \textbf{58.60} & \sbest{40.40} & \sbest{29.60} & \sbest{64.98} & \sbest{48.39} \\
 & \FT{} & 63.62 & 54.60 & 31.48 & 19.85 & 55.67 & 40.40\\ 
 \cmidrule(lr){2-8}
 & \method{} & \sbest{65.72} & \sbest{58.28} & \textbf{41.31} & \textbf{30.27} & \textbf{67.68} & \textbf{49.39} \\
 \midrule
 \multirow{2}{*}{ViT-B/32} & CLIP~\cite{radford2021learning} & 62.05 & 54.79 & 40.82 & 29.57 & \sbest{65.99} & 47.79\\
 & LP & 59.58 & 49.73 & 28.06 & 19.67 & 47.20 & 36.17 \\
 & CoOp~\cite{zhou2022learning} & \sbest{66.34} & \sbest{58.24} & \sbest{41.48} & \sbest{31.34} & 65.78 & \sbest{49.21} \\
 & \FT{} & 65.37 & 56.51 & 38.98 & 25.23 & 63.70 & 46.11 \\ \cmidrule(lr){2-8}
 & \method{} & \textbf{67.72} & \textbf{59.62} & \textbf{42.79} & \textbf{32.92} & \textbf{68.65} & \textbf{51.00}\\
\bottomrule
\end{tabular}}
\caption{Ablation studies when using different image encoders. The \textbf{best} and \sbest{second best} results on each dataset are marked.}
\label{tab:different_backbon}
 \vspace{-5pt}
\end{table}

\textbf{Visualization of important parameters.} It is instructive to ask which parts of the model contain the parameters selected by \method{} for updating. To illustrate these, we consider the CLIP model ViT-B/16, where both the image and text encoders are based on the Transformer architecture, with 12 layers for the image encoder and 16 layers for the text encoder. \cref{fig:selected_params} shows the number of learnable parameters per block of layers. Notably, \method{} tends to select more learnable parameters in the text encoder compared to the image encoder. This is consistent with our finding that fine-tuning only the text encoder is slightly better than fine-tuning only the image encoder (see \cref{tab:selection_strategy}). For the text encoder, the selected parameters are concentrated in the early layers, while for the image encoder, the selected parameters are found in the later layers. As discussed in~\cite{zhai2022lit}, textual descriptions might not be precise and clean enough to describe images. Therefore, fine-tuning early layers of the text encoder can help to describe the concepts.  Our results might provide insight into the effectiveness of prompt-tuning.

\subsection{Extension of \method{} to NLP Tasks} \label{sub:nlp}
In this section, we further validate the OOD robustness of \method{} on various NLP tasks. Experiments are carried out using pre-trained language models, namely T5-large, T5-3b~\cite{raffel2020exploring} and DeBERTa-large~\cite{he2020deberta} as backbone networks. Following Yuan \etal~\cite{yuan2023revisiting}, we use the BOSS\footnote{Available at: \url{https://github.com/lifan-yuan/OOD_NLP}} benchmark consisting of five tasks and twenty datasets. This benchmark contains a variety of NLP tasks, including sentiment analysis (SA), toxic detection (TD), and natural language inference (NLI) for classification; name entity recognition  (NER) for structured prediction; and extractive question answering (EQA) for reading comprehension. We present the best OOD performance across all tasks and datasets in~\cref{tab:boss:sa}. Please refer to~\cref{appendix:nlp} for the detailed setting and a comprehensive comparison based on various sparsity levels and also language model sizes.

We can observe that FT yields most often the best or second-best ID performance as expected. Only for TD, adapting the model with LoRA yields a better accuracy which we attribute to a possible overfitting of FT to the ID training data. \method{} consistently outperforms the conventional fine-tuning method in terms of OOD performance. Notably, \method{} surpasses LoRA in four out of five cases which empirically proves its generalizability. By only updating a few important parameters, \method{} preserves the original model's ability to generalize to new data, leading to improved average OOD performance. This demonstrates that \method{} is a model-agnostic method as well.

\begin{table}[!hb]
\tablestyle{9pt}{1.1}
\renewcommand{\arraystretch}{0.99}
\resizebox{\linewidth}{!}{
\begin{tabular}{lllrrccccc}
\toprule
\multirow{3}{*}{Task} &\multirow{3}{*}{Metric}   & \multirow{3}{*}{Method} & \multirow{3}{*}{\shortstack{\# fine-tuned\\ parameters}} & \multirow{3}{*}{Ratio} & \multicolumn{5}{c}{Datasets} \\
\cmidrule(lr){6-10}
                 &  &  &              &   & ID                    & \multicolumn{4}{c}{OOD} \\
\midrule
\multirow{5}{*}{NER} & \multirow{5}{*}{F1}  &       &                           &        & FN & CoNLL & ENER & WNUT & \textit{Average} \\
\cmidrule(lr){6-6} \cmidrule(lr){7-10}
                     &   & DeBERTa (large) & 17.43K  & 0.004\% & 52.9 & 60.0 & 36.8 & 29.5 & 42.1 \\
                     &   & FT              & 434.03M & 100\%   & \textbf{79.4} & 70.6 & 50.0 & 43.2 & 54.6 \\ 
                     &   & LoRA ($r = 8$)  & 821.28K & 0.189\% & \sbest{77.7} & 67.5 & 57.2 & 45.3 & \textbf{56.6} \\
                     &   & SAFT ($\alpha=0.00025$) &  108.57K & 0.025\% & 76.6 & 67.4 & 55.7 & 46.3 & \sbest{56.5} \\
\midrule
\multirow{5}{*}{SA} & \multirow{5}{*}{Accuracy}    &    &                           &        & AZ	& DS&	SE&	SST & \textit{Average} \\
\cmidrule(lr){6-6} \cmidrule(lr){7-10}
                      & & T5 (large)     & 0        & 0\%    & 85.78	& 35.58 & 34.55 & 43.49	& 37.87                  \\
                      & & FT             & 737.67M  & 100\%  & \textbf{91.22} & 46.69 & 46.68 & 75.07 & 56.15                  \\ 
                      & & LoRA ($r = 8$) & 2,359.30K & 0.319\% & \sbest{90.37} & 47.20 & 48.81 & 77.13 & \sbest{57.71} \\
                      & & SAFT ($\alpha=0.0005$)  & 368.83K & 0.050\% & 89.62 & 48.75 & 49.66 & 76.48 & \textbf{58.29} \\
\midrule
\multirow{5}{*}{TD} &  \multirow{5}{*}{Accuracy} &      &                           &        & CC	& AC	& IH &	TG & \textit{Average} \\
\cmidrule(lr){6-6} \cmidrule(lr){7-10}
                      &  & T5 (large)    & 0       & 0\%     & 14.90 & 79.47 & 39.54 & 43.94 & 54.31 \\
                      &  & FT            & 737.67M & 100\%   & \sbest{85.43} & 64.28 & 62.47 & 68.94 & \sbest{65.23} \\  
                      & & LoRA ($r = 4$) & 1,179.65K & 0.160\% & \textbf{86.17} & 68.29 & 60.80 & 65.21 & 64.77 \\
                      & & SAFT ($\alpha=0.0005$)  & 368.83K & 0.050\% & 85.31 & 73.15 & 61.94 & 64.89 & \textbf{66.66} \\
\midrule
\multirow{5}{*}{NLI} &   \multirow{5}{*}{Accuracy}&      &                           &        & MN &	AN	& CN &	WN & \textit{Average} \\
\cmidrule(lr){6-6} \cmidrule(lr){7-10}
                      &  & T5 (large)    & 0       & 0\%     & 35.02 & 33.03 &	45.72 &	19.22 &	32.66                  \\
                      &  & FT            & 737.67M & 100\%   & \textbf{89.25} & 37.19	& 38.79	& 62.66	&	46.21                  \\  
                      & & LoRA ($r = 8$) & 2,359.30K & 0.319\% & \sbest{89.21} & 33.44 & 46.63 & 60.26 & \sbest{46.77} \\
                      & & SAFT ($\alpha=0.00025$) & 184.42K & 0.025\% & 86.65 & 30.28 & 57.44 & 56.82 & \textbf{48.18} \\
\midrule
\multirow{5}{*}{EQA} &  \multirow{5}{*}{F1}&      &                           &        & SQuAD	& AQA	& NQA	& SQA & \textit{Average} \\
\cmidrule(lr){6-6} \cmidrule(lr){7-10}
                      &  & T5 (large)    & 0       & 0\%     & 27.41 &	10.14 &	29.96	& 21.23		& 20.45             \\
                      & & FT             & 737.67M & 100\%   & \sbest{93.36} &	49.93 &	64.36	& 38.27	&	50.85    \\ 
                      & & LoRA ($r = 4$) & 1,179.65K & 0.160\% & \textbf{93.38} & 50.88 & 66.06 & 38.99 & \sbest{51.98} \\
                      & & SAFT ($\alpha=0.00025$) & 184.42K & 0.025\% & 93.11 & 50.34 & 66.61 & 40.26 & \textbf{52.40} \\
                    \bottomrule
\end{tabular}
}
\caption{Results on the BOSS benchmark. The \textbf{best} and \sbest{second best} ID and OOD averages are marked.} 
\label{tab:boss:sa}
\vspace{-10pt}
\end{table}

\section{Related Work}
\textbf{Vision-language pre-training (VLP).} 
Recent advances have shown a remarkable improvement in VLP for downstream tasks, eliminating the need to train entirely new models from scratch. A VLP model typically consists of an image encoder, a text encoder, and an alignment mechanism for cross-modal interaction. CLIP~\cite{radford2021learning} was one of the early methods that employed contrastive learning~\cite{oord2018representation} to align image and text. ALIGN~\cite{jia2021scaling} further enhanced the scalability of vision-language representation learning on a noisy dataset containing 1.8 billion image-text pairs. More recently, BASIC~\cite{pham2023combined} combined scaling techniques that brought significant improvements to zero-shot learning tasks.

\textbf{Fine-tuning.}  While VLP models demonstrate remarkable zero-shot learning capabilities across various tasks, fine-tuning can further enhance the performance on specific downstream tasks~\cite{goyal2023finetune,wortsman2022model,Tian_2023_CVPR}.  However, fine-tuning risks overfitting on the training task, potentially reducing OOD performance. To alleviate the overfitting problem, CLIP-Adapter~\cite{gao2021clip}  introduced a residual connection to combine the adapted features with the original CLIP features. The adapter integrated new knowledge gained from the training set while retaining the prior knowledge encoded in CLIP. Tip-Adapter~\cite{zhang2021tip} constructed a key-value cache model from the few-shot training set in a non-parametric manner without the need for additional training or fine-tuning. An interesting observation by Wortsman \etal~\cite{wortsman2022robust} was that ensembling the parameters of both pre-trained and fine-tuned models could help preserving the generalization to OOD. Another idea was to maintain a temporal ensemble weighted by the Beta distribution, which combined both the pre-trained and fine-tuned models~\cite{shu2023clipood}. Following the direction of PEFT, BitFit~\cite{zaken2022bitfit} only fine-tuned the bias-terms. LoRA~\cite{hu2021lora} aimed to reduce the number of trainable parameters. In contrast to LoRA, which concentrates on the low-rank structure of the weight matrix, our focus centers on the sparsity of parameters. Furthermore, Sung \etal.~\cite{sung2021training} proposed to use the Fisher information to find sparse masks and fine-tune the model on a single downstream task. While this approach aims to reduce communication and storage costs, SAFT focuses on enhancing OOD generalization.

\textbf{Prompt-tuning.} The performance of VLP models heavily depends on the quality of the prompt design~\cite{zhou2022learning}. This is because class names themselves do not encapsulate the complete semantic information of the image, leading to inference sensitivity based on the selected words for the prompt. Moreover, handcrafted prompts may not align perfectly with what the machine finds most favorable or effective. To address the issue, Zhou \etal~\cite{zhou2022learning} introduced CoOP along with the concept of prompt-tuning. This approach refined a prompt by integrating contextual information relevant to a specific task. In particular, the context words are turned into a set of learnable continuous embeddings. During optimization, both the image encoder and text encoder are frozen, keeping the number of parameters significantly small. CoCoOP~\cite{zhou2022conditional} improved the generalization of prompt-tuning to OOD data by making the prompt conditioned on model inputs. Other approaches suggest adjusting the prompt in an unsupervised manner~\cite{huang2022unsupervised,shu2022test}, multi-modal prompt learning~\cite{khattak2023maple}, or synthesized prompts~\cite{wang2023improving}. In contrast to prompt-tuning, our method SAFT does not introduce extra parameters to the pre-trained model, simplifying the adaptation process. Furthermore, there is no additional inference latency.

\section{Conclusion and Limitation}
\textbf{Conclusion.} We have introduced \method{}, a simple task-agnostic fine-tuning technique that significantly improves the generalizability of models on OOD tasks. During fine-tuning, \method{} only updates a subset of the learnable parameters, while keeping the others frozen. We demonstrate \method{} effectiveness on multiple tasks in the VLP domain as well as in natural language. When applied to the CLIP model, \method{} outperforms other competing methods. We show the scalability of \method{} by improving the performance of T5 language model with 3B parameters. Moreover, Our comprehensive experimental results support the derived theoretical generalization bound for SAFT. 

\textbf{Limitation.} While \method{} only updates a small subset of parameters, these learnable parameters are unstructured. Accelerating unstructured sparsity can be challenging on hardware that is primarily optimized for dense computations~\cite{hooker2021hardware}. However, recent advancements in hardware, like Procrustes~\cite{9251866} and Cerebras CS-2~\cite{10123162}, and algorithms such as SparseAdam~\cite{paszke2017automatic} and SparseProp~\cite{nikdan2023sparseprop} offer promising solutions. While our current work does not implement these optimizations, their potential for enhancing \method{}'s efficiency is notable. Another direction is to extend \method{} with learnable parameters being more structured.


%
%
\bibliographystyle{splncs04}
\bibliography{main}

\begin{thebibliography}{10}
\providecommand{\url}[1]{\texttt{#1}}
\providecommand{\urlprefix}{URL }
\providecommand{\doi}[1]{https://doi.org/#1}

\bibitem{aghajanyan2020intrinsic}
Aghajanyan, A., Zettlemoyer, L., Gupta, S.: Intrinsic dimensionality explains the effectiveness of language model fine-tuning. In: ACL. pp. 7319--7328 (2020)

\bibitem{arora2018stronger}
Arora, S., Ge, R., Neyshabur, B., Zhang, Y.: Stronger generalization bounds for deep nets via a compression approach. In: ICML. pp. 254--263 (2018)

\bibitem{bossard2014food}
Bossard, L., Guillaumin, M., Van~Gool, L.: Food-101--mining discriminative components with random forests. In: ECCV. pp. 446--461 (2014)

\bibitem{bromley1993signature}
Bromley, J., Guyon, I., LeCun, Y., S{\"a}ckinger, E., Shah, R.: Signature verification using a" siamese" time delay neural network. In: NIPS. pp. 737--739 (1993)

\bibitem{cheng2023survey}
Cheng, H., Zhang, M., Shi, J.Q.: A survey on deep neural network pruning-taxonomy, comparison, analysis, and recommendations. arXiv preprint arXiv:2308.06767  (2023)

\bibitem{cimpoi2014describing}
Cimpoi, M., Maji, S., Kokkinos, I., Mohamed, S., Vedaldi, A.: Describing textures in the wild. In: CVPR. pp. 3606--3613 (2014)

\bibitem{deng2009imagenet}
Deng, J., Dong, W., Socher, R., Li, L.J., Li, K., Fei-Fei, L.: Imagenet: A large-scale hierarchical image database. In: CVPR. pp. 248--255 (2009)

\bibitem{dosovitskiy2020image}
Dosovitskiy, A., Beyer, L., Kolesnikov, A., Weissenborn, D., Zhai, X., Unterthiner, T., Dehghani, M., Minderer, M., Heigold, G., Gelly, S., et~al.: An image is worth 16x16 words: Transformers for image recognition at scale. In: ICLR (2021)

\bibitem{fei2004learning}
Fei-Fei, L., Fergus, R., Perona, P.: Learning generative visual models from few training examples: An incremental bayesian approach tested on 101 object categories. In: CVPR workshop. pp. 178--178 (2004)

\bibitem{frankle2018lottery}
Frankle, J., Carbin, M.: The lottery ticket hypothesis: Finding sparse, trainable neural networks. In: ICLR (2018)

\bibitem{fu2023effectiveness}
Fu, Z., Yang, H., So, A.M.C., Lam, W., Bing, L., Collier, N.: On the effectiveness of parameter-efficient fine-tuning. In: AAAI. pp. 12799--12807 (2023)

\bibitem{gao2021clip}
Gao, P., Geng, S., Zhang, R., Ma, T., Fang, R., Zhang, Y., Li, H., Qiao, Y.: Clip-adapter: Better vision-language models with feature adapters. IJCV pp. 1--15 (2023)

\bibitem{goyal2023finetune}
Goyal, S., Kumar, A., Garg, S., Kolter, Z., Raghunathan, A.: Finetune like you pretrain: Improved finetuning of zero-shot vision models. In: CVPR. pp. 19338--19347 (2023)

\bibitem{guo2021parameter}
Guo, D., Rush, A.M., Kim, Y.: Parameter-efficient transfer learning with diff pruning. In: ACL. pp. 4884--4896 (2021)

\bibitem{han2015learning}
Han, S., Pool, J., Tran, J., Dally, W.: Learning both weights and connections for efficient neural network. In: NIPS (2015)

\bibitem{he2016deep}
He, K., Zhang, X., Ren, S., Sun, J.: Deep residual learning for image recognition. In: CVPR. pp. 770--778 (2016)

\bibitem{he2020deberta}
He, P., Liu, X., Gao, J., Chen, W.: Deberta: Decoding-enhanced bert with disentangled attention. In: ICLR (2021)

\bibitem{helber2019eurosat}
Helber, P., Bischke, B., Dengel, A., Borth, D.: Eurosat: A novel dataset and deep learning benchmark for land use and land cover classification. IEEE J. Sel. Top. Appl. Earth Obs.  \textbf{12},  2217--2226 (2019)

\bibitem{hendrycks2021many}
Hendrycks, D., Basart, S., Mu, N., Kadavath, S., Wang, F., Dorundo, E., Desai, R., Zhu, T., Parajuli, S., Guo, M., et~al.: The many faces of robustness: A critical analysis of out-of-distribution generalization. In: CVPR. pp. 8340--8349 (2021)

\bibitem{hendrycks2021natural}
Hendrycks, D., Zhao, K., Basart, S., Steinhardt, J., Song, D.: Natural adversarial examples. In: CVPR. pp. 15262--15271 (2021)

\bibitem{hooker2021hardware}
Hooker, S.: The hardware lottery. Communications of the ACM  \textbf{64}(12),  58--65 (2021)

\bibitem{houlsby2019parameter}
Houlsby, N., Giurgiu, A., Jastrzebski, S., Morrone, B., De~Laroussilhe, Q., Gesmundo, A., Attariyan, M., Gelly, S.: Parameter-efficient transfer learning for nlp. In: ICML. pp. 2790--2799 (2019)

\bibitem{hu2021lora}
Hu, E.J., Shen, Y., Wallis, P., Allen-Zhu, Z., Li, Y., Wang, S., Wang, L., Chen, W.: Lora: Low-rank adaptation of large language models. In: ICLR (2021)

\bibitem{huang2022unsupervised}
Huang, T., Chu, J., Wei, F.: Unsupervised prompt learning for vision-language models. arXiv preprint arXiv:2204.03649  (2022)

\bibitem{jia2021scaling}
Jia, C., Yang, Y., Xia, Y., Chen, Y.T., Parekh, Z., Pham, H., Le, Q., Sung, Y.H., Li, Z., Duerig, T.: Scaling up visual and vision-language representation learning with noisy text supervision. In: ICML. pp. 4904--4916 (2021)

\bibitem{khattak2023maple}
Khattak, M.U., Rasheed, H., Maaz, M., Khan, S., Khan, F.S.: Maple: Multi-modal prompt learning. In: CVPR. pp. 19113--19122 (2023)

\bibitem{kirkpatrick2017overcoming}
Kirkpatrick, J., Pascanu, R., Rabinowitz, N., Veness, J., Desjardins, G., Rusu, A.A., Milan, K., Quan, J., Ramalho, T., Grabska-Barwinska, A., et~al.: Overcoming catastrophic forgetting in neural networks. Proceedings of the national academy of sciences  \textbf{114},  3521--3526 (2017)

\bibitem{kornblith2019better}
Kornblith, S., Shlens, J., Le, Q.V.: Do better imagenet models transfer better? In: CVPR. pp. 2661--2671 (2019)

\bibitem{krause20133d}
Krause, J., Stark, M., Deng, J., Fei-Fei, L.: 3d object representations for fine-grained categorization. In: CVPR workshops. pp. 554--561 (2013)

\bibitem{kumar2022fine}
Kumar, A., Raghunathan, A., Jones, R., Ma, T., Liang, P.: Fine-tuning can distort pretrained features and underperform out-of-distribution. In: ICLR (2022)

\bibitem{li2018measuring}
Li, C., Farkhoor, H., Liu, R., Yosinski, J.: Measuring the intrinsic dimension of objective landscapes. In: ICLR (2018)

\bibitem{10123162}
Lie, S.: Cerebras architecture deep dive: First look inside the hardware/software co-design for deep learning. IEEE Micro  \textbf{43},  18--30 (2023)

\bibitem{loshchilov2017decoupled}
Loshchilov, I., Hutter, F.: Decoupled weight decay regularization. In: ICLR (2017)

\bibitem{maji2013fine}
Maji, S., Rahtu, E., Kannala, J., Blaschko, M., Vedaldi, A.: Fine-grained visual classification of aircraft. arXiv preprint arXiv:1306.5151  (2013)

\bibitem{miller2021accuracy}
Miller, J.P., Taori, R., Raghunathan, A., Sagawa, S., Koh, P.W., Shankar, V., Liang, P., Carmon, Y., Schmidt, L.: Accuracy on the line: on the strong correlation between out-of-distribution and in-distribution generalization. In: ICML. pp. 7721--7735 (2021)

\bibitem{nikdan2023sparseprop}
Nikdan, M., Pegolotti, T., Iofinova, E., Kurtic, E., Alistarh, D.: Sparseprop: Efficient sparse backpropagation for faster training of neural networks at the edge. In: ICML (2023)

\bibitem{nilsback2008automated}
Nilsback, M.E., Zisserman, A.: Automated flower classification over a large number of classes. In: ICVGIP. pp. 722--729 (2008)

\bibitem{oord2018representation}
Oord, A.v.d., Li, Y., Vinyals, O.: Representation learning with contrastive predictive coding. arXiv preprint arXiv:1807.03748  (2018)

\bibitem{parkhi2012cats}
Parkhi, O., Vedaldi, A., Jawahar, C., Zisserman, A.: Cats and dogs. In: CVPR. p. 3498–3505 (2012)

\bibitem{paszke2017automatic}
Paszke, A., Gross, S., Chintala, S., Chanan, G., Yang, E., DeVito, Z., Lin, Z., Desmaison, A., Antiga, L., Lerer, A.: Automatic differentiation in pytorch. In: NIPS-W (2017)

\bibitem{pham2023combined}
Pham, H., Dai, Z., Ghiasi, G., Kawaguchi, K., Liu, H., Yu, A.W., Yu, J., Chen, Y.T., Luong, M.T., Wu, Y., et~al.: Combined scaling for zero-shot transfer learning. Neurocomputing  (2023)

\bibitem{radford2021learning}
Radford, A., Kim, J.W., Hallacy, C., Ramesh, A., Goh, G., Agarwal, S., Sastry, G., Askell, A., Mishkin, P., Clark, J., et~al.: Learning transferable visual models from natural language supervision. In: ICML. pp. 8748--8763 (2021)

\bibitem{raffel2020exploring}
Raffel, C., Shazeer, N., Roberts, A., Lee, K., Narang, S., Matena, M., Zhou, Y., Li, W., Liu, P.J.: Exploring the limits of transfer learning with a unified text-to-text transformer. JMLR  \textbf{21},  5485--5551 (2020)

\bibitem{recht2019imagenet}
Recht, B., Roelofs, R., Schmidt, L., Shankar, V.: Do imagenet classifiers generalize to imagenet? In: ICML. pp. 5389--5400 (2019)

\bibitem{russakovsky2015imagenet}
Russakovsky, O., Deng, J., Su, H., Krause, J., Satheesh, S., Ma, S., Huang, Z., Karpathy, A., Khosla, A., Bernstein, M., et~al.: Imagenet large scale visual recognition challenge. IJCV  \textbf{115},  211--252 (2015)

\bibitem{shu2022test}
Shu, M., Nie, W., Huang, D.A., Yu, Z., Goldstein, T., Anandkumar, A., Xiao, C.: Test-time prompt tuning for zero-shot generalization in vision-language models. In: NIPS. pp. 14274--14289 (2022)

\bibitem{shu2023clipood}
Shu, Y., Guo, X., Wu, J., Wang, X., Wang, J., Long, M.: Clipood: Generalizing clip to out-of-distributions. In: ICML. pp. 31716--31731 (2023)

\bibitem{soomro2012ucf101}
Soomro, K., Zamir, A.R., Shah, M.: Ucf101: A dataset of 101 human actions classes from videos in the wild. arXiv preprint arXiv:1212.0402  (2012)

\bibitem{sung2021training}
Sung, Y.L., Nair, V., Raffel, C.A.: Training neural networks with fixed sparse masks. In: NeurIPS. vol.~34, pp. 24193--24205 (2021)

\bibitem{Tian_2023_CVPR}
Tian, J., He, Z., Dai, X., Ma, C.Y., Liu, Y.C., Kira, Z.: Trainable projected gradient method for robust fine-tuning. In: Proceedings of the IEEE/CVF Conference on Computer Vision and Pattern Recognition (CVPR). pp. 7836--7845 (June 2023)

\bibitem{vaswani2017attention}
Vaswani, A., Shazeer, N., Parmar, N., Uszkoreit, J., Jones, L., Gomez, A.N., Kaiser, {\L}., Polosukhin, I.: Attention is all you need. In: NIPS (2017)

\bibitem{wang2019learning}
Wang, H., Ge, S., Lipton, Z., Xing, E.P.: Learning robust global representations by penalizing local predictive power. In: NIPS (2019)

\bibitem{wang2023improving}
Wang, Z., Liang, J., He, R., Xu, N., Wang, Z., Tan, T.: Improving zero-shot generalization for clip with synthesized prompts. In: ICCV. pp. 3032--3042 (2023)

\bibitem{wortsman2022model}
Wortsman, M., Ilharco, G., Gadre, S.Y., Roelofs, R., Gontijo-Lopes, R., Morcos, A.S., Namkoong, H., Farhadi, A., Carmon, Y., Kornblith, S., et~al.: Model soups: averaging weights of multiple fine-tuned models improves accuracy without increasing inference time. In: ICML. pp. 23965--23998 (2022)

\bibitem{wortsman2022robust}
Wortsman, M., Ilharco, G., Kim, J.W., Li, M., Kornblith, S., Roelofs, R., Lopes, R.G., Hajishirzi, H., Farhadi, A., Namkoong, H., et~al.: Robust fine-tuning of zero-shot models. In: CVPR. pp. 7959--7971 (2022)

\bibitem{xian2017zero}
Xian, Y., Schiele, B., Akata, Z.: Zero-shot learning-the good, the bad and the ugly. In: CVPR. pp. 4582--4591 (2017)

\bibitem{xiao2010sun}
Xiao, J., Hays, J., Ehinger, K.A., Oliva, A., Torralba, A.: Sun database: Large-scale scene recognition from abbey to zoo. In: CVPR. pp. 3485--3492 (2010)

\bibitem{9251866}
Yang, D., Ghasemazar, A., Ren, X., Golub, M., Lemieux, G., Lis, M.: Procrustes: a dataflow and accelerator for sparse deep neural network training. In: Proceedings of the Annual IEEE/ACM International Symposium on Microarchitecture. pp. 711--724 (2020)

\bibitem{yuan2023revisiting}
Yuan, L., Chen, Y., Cui, G., Gao, H., Zou, F., Cheng, X., Ji, H., Liu, Z., Sun, M.: Revisiting out-of-distribution robustness in nlp: Benchmark, analysis, and llms evaluations. In: NeurIPS D\&B Track (2023)

\bibitem{zaken2022bitfit}
Zaken, E.B., Goldberg, Y., Ravfogel, S.: Bitfit: Simple parameter-efficient fine-tuning for transformer-based masked language-models. In: ACL. pp.~1--9 (2022)

\bibitem{zhai2022lit}
Zhai, X., Wang, X., Mustafa, B., Steiner, A., Keysers, D., Kolesnikov, A., Beyer, L.: Lit: Zero-shot transfer with locked-image text tuning. In: CVPR. pp. 18123--18133 (2022)

\bibitem{zhang2021can}
Zhang, D., Ahuja, K., Xu, Y., Wang, Y., Courville, A.: Can subnetwork structure be the key to out-of-distribution generalization? In: ICML. pp. 12356--12367 (2021)

\bibitem{zhang2021tip}
Zhang, R., Fang, R., Zhang, W., Gao, P., Li, K., Dai, J., Qiao, Y., Li, H.: Tip-adapter: Training-free clip-adapter for better vision-language modeling. arXiv preprint arXiv:2111.03930  (2021)

\bibitem{zhou2022conditional}
Zhou, K., Yang, J., Loy, C.C., Liu, Z.: Conditional prompt learning for vision-language models. In: CVPR. pp. 16816--16825 (2022)

\bibitem{zhou2022learning}
Zhou, K., Yang, J., Loy, C.C., Liu, Z.: Learning to prompt for vision-language models. IJCV  \textbf{130},  2337--2348 (2022)

\end{thebibliography}
\appendix
\clearpage
\setcounter{page}{1}


\section{Generalization Bound} \label{sub:bound}
We provide an asymptotic generalization bound for \method{} in ID settings. Our result is based on a compression framework introduced by Arora \etal~\cite{arora2018stronger}. A similar analysis has been carried out in~\cite{aghajanyan2020intrinsic}. More specifically, consider a multi-class classification problem where the expected margin loss is defined as
\begin{align}
    \mathcal{L}_{\gamma}(f_\theta) = \mathbb{E}_{(\rvx,\ry)} \big[ f_\theta(\vx, y) \le \gamma + \max_{c\ne y} f_\theta(\vx, c)\big] 
\end{align}
with $\gamma \ge 0$ a pre-defined margin. The standard classification loss corresponds to $\gamma=0$. Let  $\hat{\mathcal{L}}_\gamma$ denote an empirical estimate of the margin loss. We define the parameters of the fine-tuned model as $\tilde{\theta} = \theta + \vm \odot \Delta\theta$, where $\Delta\theta$ denotes a difference vector.
\begin{thm} \label{theorem:bound}
Let $\mathcal{G} = \{ f_{\tilde{\theta}} \mid \tilde{\theta} \in \tilde{\Theta} \}$ be a set of classifiers $f_{\tilde{\theta}}$, where $\tilde{\theta}$ consists of $d$ parameters each of which can have at most $r$ discrete values. Given a dataset of $N$ examples, there exists an $\tilde{\theta} \in \tilde{\Theta}$ over the training dataset with a high probability such that
\begin{align}
    \mathcal{L}_0 (f_{\tilde{\theta}}) \le \hat{\mathcal{L}}_0(f_\theta) + O\left( \sqrt{\frac{d\log r}{N}} \right)\,.
\end{align}
\vspace{-10pt}
\end{thm}
Theorem~\ref{theorem:bound} demonstrates that the in-domain generalization error does not depend on the number of parameters. Instead, it hinges on the number of trainable parameters that are used for adapting to the downstream task. The theorem shows that \method{} can effectively narrow the in-domain generalization error compared to conventional fine-tuning. In other words, there exists a small set of parameters that is as effective for fine-tuning as the full parameter space. While this theorem only covers ID generalization, it is still meaningful as the OOD error can be understood as a combination of ID and OOD generalization. 

\textbf{Proof of Theorem~\ref{theorem:bound}.} For the sake of completeness, we revisit the definition of $(\gamma, S)$-compressible using helper string $s$ introduced by Arora \etal~\cite{arora2018stronger}.
\begin{defi}[$(\gamma, S)$-compressible using helper string $s$]~\cite{arora2018stronger}
Suppose $G_{\mathcal{A},s} =\{g_{A,s}|A\in \mathcal{A}\}$ is a class of classifiers indexed by trainable parameters $A$ and fixed strings $s$.  A classifier $f$ is  ($\gamma,S$)-compressible with respect to $G_{\mathcal{A},s}$ using helper string $s$ if there exists $A\in \mathcal{A}$ such that for any $x\in S$, we have for all $y$
\begin{equation*}
	|f(x)[y] - g_{A,s}(x)[y]| \le \gamma.
\end{equation*}
\end{defi}

It is straightforward to see that \method{} is  $(0, S)$-compressible with a helper string being a random seed used to generate the learnable parameters in $\tilde{\theta}$ and initial pre-trained model parameters for frozen parameters. By setting the learnable parameters to the original pre-trained parameters $f$ is losslessly compressible.

Therefore, our generalization bound directly follows from Theorem 2.1 in \cite{arora2018stronger},
\begin{align}
    \mathcal{L}_0 (f_{\tilde{\theta}}) \le \hat{\mathcal{L}}_0(f_\theta) + O\left( \sqrt{\frac{d\log r}{N}} \right)\,.
\end{align}
Note that the parameters $\tilde{\theta}$  can be discretized using quantization. As done in~\cite{aghajanyan2020intrinsic}, the number of discrete states $r$ depends on the level of quantization (FP32 or FP16). 

\section{Experimental Details}
\subsection{Datasets}  \label{appendix:dataset}
In this section, we provide more details about the datasets used in our experiments.  Table \ref{table:datasets} summarizes the train, validation, and test set along with the prompt template\footnote{All data splits are available at \url{https://github.com/KaiyangZhou/CoOp/blob/main/DATASETS.md}}. Next, we provide a brief overview of each dataset.

\textbf{ImageNet}~\cite{deng2009imagenet} is a large-scale visual dataset containing over 1 million labeled images spanning thousands of object categories. It has been widely used for training and evaluating computer vision models, particularly for image classification tasks.

\textbf{ImageNet-V2}~\cite{recht2019imagenet} contains 10K test images gathered from the Flicker image hosting service. To make sure that the new set of images follows the same distribution as the original ImageNet~\cite{deng2009imagenet} dataset, only those images uploaded in a similar time frame as ImageNet have been considered.

\textbf{ImageNet-Sketch}~\cite{wang2019learning} consists of 50k images, with 50 images for each of 1000 ImageNet classes. It was created by collecting results from 100 Google Image queries for each class and then manually filtering out irrelevant images. There are some classes with less than 50 samples after cleaning that are compensated by augmenting the dataset by flipping and rotating the images.

\textbf{ImageNet-A}~\cite{hendrycks2021natural} consists of 7,500 adversarially filtered examples of ImageNet, which often cause low performance for classifiers. It only contains 200 classes selected from the original set of 1000 ImageNet classes.

\textbf{ImgeNet-R}~\cite{hendrycks2021many} contains 30k images gathered based on renditions of 200 ImageNet classes. The renditions are in the form of art, cartoons, deviantart, graffiti, embroidery, graphics, origami, paintings, patterns, plastic objects, plush objects, sculptures, sketches, tattoos, toys, and video games.

\textbf{Caltech101}~\cite{fei2004learning} contains approximately 9k images of objects distributed across 101 categories. The categories contain varying numbers of images, ranging from 40 to 800 images per category. Most categories have around 50 images, each with a size of $300 \times 200$ pixels.

\textbf{OxfordPets}~\cite{parkhi2012cats} is a dataset that includes 7,349 images distributed among 37 categories of cats and dogs. Each class contains approximately 200 images. The dataset exhibits significant variations in scale, pose, and lighting.

\textbf{StanfordCars}~\cite{krause20133d} consists of 16,185 images of 196 classes. The images have a size of $360 \times 240$ pixels. Categories are typically at the level of Make, Model, and Year. 

\textbf{Flowers102}~\cite{nilsback2008automated} contains over 8,000 images distributed across 102 flower categories. Each class has between 40 and 258 images, with large variations within a class.

\textbf{Food101}~\cite{bossard2014food} consists of 101k images of 101 categories with 250 manually reviewed test images in addition to 750 training images. The training data are noisy in the form of intense colors and wrong labels. The images are re-scaled such that their maximum side length is 512 pixels.

\textbf{FGVCAircraft}~\cite{maji2013fine} is designed for fine-grained visual categorization of aircraft. There are 10,200 images with 100 images for each of 102 different aircraft model variants, most of which are airplanes.

\textbf{SUN397}~\cite{xiao2010sun} consists of 16,873 images of 397  categories, used in the Scene UNderstanding (SUN) benchmark to evaluate algorithms for scene recognition.

\textbf{UCF101}~\cite{soomro2012ucf101} is an action recognition dataset containing 13,320 videos of 101 actions collected from YouTube. The videos are split into training, validation, and test sets. We use the middle frame of each video as image input.

\textbf{DTD}~\cite{cimpoi2014describing} consists of 5,640 images of 47 categories each representing a describable texture from a human perceptual perspective. There are 120 images for each category in the dataset.

\textbf{EuroSAT}~\cite{helber2019eurosat} consists of 27k Sentinel-2 satellite images of 10 classes that cover 13 spectral bands. This dataset is used for the task of satellite image classification.

\begin{table*}[t!]
\centering
\resizebox{\linewidth}{!}{
\begin{tabular}{lrrrrl}
\toprule
Dataset & Classes & Train &Val & Test & Prompt\\
\midrule
ImageNet & 1k & 1.28M &N/A & 50k & ``a photo of a [CLASS].''\\
ImageNet V2 & 1k & N/A&N/A & 10k & ``a photo of a [CLASS].''\\
ImageNet-Sketch& 1k & N/A&N/A & 50889 & ``a photo of a [CLASS].''\\
ImageNet-A& 200 & N/A&N/A & 7500 & ``a photo of a [CLASS].''\\
ImageNet-R& 200 & N/A&N/A & 30k & ``a photo of a [CLASS].''\\
\midrule
Caltech101 & 100 & 4128 &1649 & 2465 & ``a photo of a [CLASS].''\\
OxfordPets & 37 & 2944 &736 & 3669 & ``a photo of a [CLASS], a type of pet.''\\
StanfordCars & 196 & 6509 & 1635 & 8041 & ``a photo of a [CLASS].''\\
Flowers102 & 102 & 4093 & 1633 & 2463 & ``a photo of a [CLASS], a type of flower.''\\
Food101 & 101 &  50500 & 20200  & 30300 & ``a photo of a [CLASS], a type of food.''\\
FGVCAircraft & 100 &   3334  & 3333  & 3333 & ``a photo of a [CLASS], a type of aircraft.''\\
SUN397  & 397 & 15880 & 3970 & 19850 & ``a photo of a [CLASS].''\\
DTD  & 47 & 2820 & 1128 & 1692 & ``[CLASS] texture.''\\
EuroSAT  & 10 &  13500 & 5400 & 8100 & ``a centered satellite photo of [CLASS].''\\
UCF101  & 101 &  7639 & 1898 & 3783 & ``a photo of a person doing [CLASS].''\\
\bottomrule
\end{tabular}}
\caption{Summary of datasets used in our experiments} \label{table:datasets}
\end{table*}

\subsection{Baseline Methods} \label{appendix:baseline}

In this section, we provide more details about the competing methods used in our experiments. 

\textbf{WiSE-FT}~\cite{wortsman2022robust} is a simple yet effective fine-tuning method that alleviates the catastrophic forgetting problem during fine-tuning by taking the linear combination of pre-trained model parameters with the fine-tuned counterpart. It has been empirically shown that WiSE-FT can preserve the generalization to OOD data.


\textbf{CLIPood}~\cite{shu2023clipood} aims to improve OOD generalization of the CLIP model on downstream tasks by incorporating the textual knowledge of the text encoder. More specifically, CLIPood uses the Margin Metric Softmax (MMS) loss during the fine-tuning phase to consider semantic relations among class names and image embeddings. The distance between the correct class label and other classes acts like an adaptive margin that pushes the image representation further apart from incorrect class labels. Moreover, to ensure that during fine-tuning, the model preserves its pre-trained knowledge while customized to a new dataset, Beta Temporal Ensemble is used. 

\textbf{LoRA}~\cite{hu2021lora} addresses the compute and memory requirements of large language models during fine-tuning for downstream tasks. The main intuition is that the update to the weights resides in a low-dimensional subspace of the parameter space. The update matrix can be represented as a low-rank decomposition of two matrices. 

\textbf{CoOp}~\cite{zhou2022learning}  introduces learnable context vectors to replace hand-crafted prompt engineering for vision and language models, such as CLIP. These vectors are trained end-to-end through fine-tuning, minimizing the classification loss for these vectors while keeping the rest of the parameters frozen. Although CoOp demonstrates significant improvement over hand-crafted prompts, its generalizability to unseen classes is limited. 

\textbf{CoCoOp}~\cite{zhou2022conditional} is an extension of CoOp that improves generalizability by introducing dynamic context tokens conditioned on input images. It achieves this by training a meta-network along with context vectors to generate conditional context tokens for each image. While the dynamic context generation improves generalizability compared to CoOp, CoCoOp is notably slower than its predecessor.

\textbf{MaPLe}~\cite{khattak2023maple} extends the concept of prompt learning to both vision and text modalities. In particular, each transformer block has a set of learnable prompts, and the vision prompts are explicitly conditioned on their textual prompt counterparts through coupling blocks. 


\section{Further Experiments} 
\subsection{Detailed Results  from Base to New Classes} \label{appendix:base_to_new}
We provide the results of generalization from base to new classes as in \cref{sub:generalization_new_classes}. The results on each dataset and the average results over 11 datasets are shown in Table~\ref{tab:on_new_classes_benchmark}. On most datasets, \method{} consistently ranks as the best or second-best method. \method{} not only improves the performance in downstream tasks but also maintains its generalization capabilities for new classes.

\begin{table}[t]
    \small
    
    \begin{subtable}[t]{.3\linewidth}
    \centering
    \resizebox{\linewidth}{!}{\begin{tabular}{lccc}
    \toprule
    & Base & New & H \\
    \midrule
    CLIP & 69.34 & 74.22 & 71.70 \\
    CoOp & 82.69 & 63.22 & 71.66 \\
    CoCoOp & 80.47 & 71.69 & 75.83 \\
    CLIPood & 83.90 & 74.50 & \sbest{78.90} \\
    MaPLe & 82.28 & \textbf{75.14} & 78.55 \\
    \FT{} & \textbf{85.86} & 71.23& 77.86 \\
    \midrule
    \method{} & \sbest{83.97} &  \sbest{74.78} & \textbf{79.11} \\
    \bottomrule
    \end{tabular}}
    \caption{Average over 11 datasets}
    \end{subtable}
    \hfill
    \begin{subtable}[t]{.3\textwidth}
    \centering
    \resizebox{\linewidth}{!}{\begin{tabular}{lccc}
    \toprule
    & Base & New & H \\
    \midrule
    CLIP & 72.43 & 68.14 & 70.22 \\
    CoOp & 76.47 & 67.88 & 71.92\\
    CoCoOp & 75.98 & 70.43 & 73.10 \\
    CLIPood & \sbest{77.50} & 70.30 & \sbest{73.70} \\
    MaPLe & 76.66 & \sbest{70.54} & 73.47 \\
    \FT{} & 75.27 & 68.29 & 71.61 \\
    \midrule
    \method{} & \textbf{77.74} & \textbf{71.40} & \textbf{74.44} \\
    \bottomrule
    \end{tabular}}
    \caption{ImageNet}
    \end{subtable}
    \hfill
    \begin{subtable}[t]{.3\textwidth}
    \centering
    \resizebox{\linewidth}{!}{\begin{tabular}{lccc}
    \toprule
    & Base & New & H \\
    \midrule
    CLIP & 96.84 & 94.00 & 95.40 \\
    CoOp & 98.00 & 89.81 & 93.73 \\
    CoCoOp & 97.96 & 93.81 & 95.84 \\
    CLIPood & \textbf{98.70} & \textbf{94.60} & \textbf{96.60} \\
    MaPLe & 97.74 & 94.36 & 96.02 \\
    \FT{} & \sbest{98.64} & 93.23 & 95.86 \\
    \midrule
    \method{} & 98.06 & \sbest{94.54} & \sbest{96.27} \\
    \bottomrule
    \end{tabular}}
    \caption{Caltech101}
    \end{subtable}
    \vspace{1em}
    \hfill
    
    \begin{subtable}[t]{.3\textwidth}
    \centering
    \resizebox{\linewidth}{!}{\begin{tabular}{lccc}
    \toprule
    & Base & New & H \\
    \midrule
    CLIP & 91.17 & 97.26 & 94.12 \\
    CoOp & 93.67 & 95.29 & 94.47 \\
    CoCoOp & 95.20 & 97.69 & \sbest{96.43} \\
    CLIPood & \textbf{95.70} & 96.40 & 96.00\\
    MaPLe & \sbest{95.43} & \textbf{97.76} & \textbf{96.58} \\
    \FT{} & 95.22 & 97.54 & 96.37 \\
    \midrule
    \method{} & 95.06 & \sbest{97.71} & 96.37 \\
    \bottomrule
    \end{tabular}}
    \caption{OxfordPets}
    \end{subtable}
    \hfill
    \begin{subtable}[t]{.3\textwidth}
    \centering
    \resizebox{\linewidth}{!}{\begin{tabular}{lccc}
    \toprule
    & Base & New & H \\
    \midrule
    CLIP & 63.37 & \textbf{74.89} & 68.65 \\
    CoOp & 78.12 & 60.40 & 68.13 \\
    CoCoOp & 70.49 & 73.59 & 72.01 \\
    CLIPood & \sbest{78.60} & 73.50 & \sbest{75.90} \\
    MaPLe & 72.94 & \sbest{74.00} & 73.47 \\
    \FT{} & 84.18 & 65.44 & 73.64\\
    \midrule
    \method{} & \textbf{78.41} & 73.61 & \textbf{75.93} \\
    \bottomrule
    \end{tabular}}
    \caption{StanfordCars}
    \end{subtable}
    \hfill
    \begin{subtable}[t]{.3\textwidth}
    \centering
    \resizebox{\linewidth}{!}{\begin{tabular}{lccc}
    \toprule
    & Base & New & H \\
    \midrule
    CLIP & 72.08 & \textbf{77.80} & 74.83 \\
    CoOp & \sbest{97.60} & 59.67 & 74.06 \\
    CoCoOp & 94.87 & 71.75 & 81.71 \\
    CLIPood & 93.50 & 74.50 & \sbest{82.90} \\
    MaPLe & 95.92 & 72.46 & 82.56 \\
    \FT{} & \textbf{98.39} & 70.78 & 82.33 \\
    \midrule
    \method{} & 96.77 & \sbest{75.46} & \textbf{84.80}\\
    \bottomrule
    \end{tabular}}
    \caption{Flowers102}
    \end{subtable}
    \vspace{1em}
    \hfill
    
    \begin{subtable}[t]{.3\textwidth}
    \centering
    \resizebox{\linewidth}{!}{\begin{tabular}{lccc}
    \toprule
    & Base & New & H \\
    \midrule
    CLIP & 90.10 & 91.22 & 90.66 \\
    CoOp & 88.33 & 82.26 & 85.19 \\
    CoCoOp & \sbest{90.70} & 91.29 & 90.99 \\
    CLIPood & \sbest{90.70} & \sbest{91.70} & \sbest{91.20}  \\
    MaPLe & \textbf{90.71} & \textbf{92.05} & \textbf{91.38} \\
    \FT{} & 86.84 & 88.69 & 87.76\\
    \midrule
    \method{} &  90.03 &  91.51 & 90.76 \\
    \bottomrule
    \end{tabular}}
    \caption{Food101}
    \end{subtable}
    \hfill
    \begin{subtable}[t]{.3\textwidth}
    \centering
    \resizebox{\linewidth}{!}{\begin{tabular}{lccc}
    \toprule
    & Base & New & H \\
    \midrule
    CLIP & 27.19 & 36.29 & 31.09 \\
    CoOp & 40.44 & 22.30 & 28.75 \\
    CoCoOp & 33.41 & 23.71 & 27.74 \\
    CLIPood & 43.30 & \textbf{37.20} &\textbf{40.00}  \\
    MaPLe & 37.44 & \sbest{35.61} & 36.50 \\
    \FT{} & \textbf{55.94} & 23.88 & 33.47  \\
    \midrule
    \method{} & \sbest{44.54} & 32.15 & \sbest{37.34} \\
    \bottomrule
    \end{tabular}}
    \caption{FGVCAircraft}
    \end{subtable}
    \hfill
    \begin{subtable}[t]{.3\textwidth}
    \centering
    \resizebox{\linewidth}{!}{\begin{tabular}{lccc}
    \toprule
    & Base & New & H \\
    \midrule
    CLIP & 69.36 & 75.35 & 72.23 \\
    CoOp & 80.60 & 65.89 & 72.51 \\
    CoCoOp & 79.74 & 76.86 & 78.27 \\
    CLIPood & 81.00 & \sbest{79.30} & \sbest{80.20}  \\
    MaPLe & 80.82 & 78.70 & 79.75 \\
    \FT{} &  \sbest{81.07} & 76.82 & 78.89 \\
    \midrule
    \method{} & \textbf{81.92} & \textbf{79.35} & \textbf{80.61} \\
    \bottomrule
    \end{tabular}}
    \caption{SUN397}
    \end{subtable}
    \vspace{1em}

    \begin{subtable}[t]{.3\textwidth}
    \centering
    \resizebox{\linewidth}{!}{\begin{tabular}{lccc}
    \toprule
    & Base & New & H \\
    \midrule
    CLIP & 53.24 & 59.90 & 56.37 \\
    CoOp & 79.44 & 41.18 & 54.24 \\
    CoCoOp & 77.01 & 56.00 & 64.85 \\
    CLIPood & \sbest{80.80} & 58.60 & 67.90  \\
    MaPLe & 80.36 & 59.18 & 68.16 \\
    \FT{} & \textbf{84.49} & \sbest{62.92} & \textbf{72.13} \\
    \midrule
    \method{} & 79.86 & \textbf{64.73} & \sbest{71.50} \\
    \bottomrule
    \end{tabular}}
    \caption{DTD}
    \end{subtable}
    \hfill
    \begin{subtable}[t]{.3\textwidth}
    \centering
    \resizebox{\linewidth}{!}{\begin{tabular}{lccc}
    \toprule
    & Base & New & H \\
    \midrule
    CLIP & 56.48 & 64.05 & 60.03 \\
    CoOp & 92.19 & 54.74 & 68.69 \\
    CoCoOp & 87.49 & 60.04 & 71.21 \\
    CLIPood & \textbf{97.50} & \sbest{64.10} & \sbest{77.30}   \\
    MaPLe & 94.07 & \textbf{73.23} & \textbf{82.35} \\
    \FT{} & \sbest{97.00} & 57.90 &  72.52  \\
    \midrule
    \method{} & 96.12 & 63.72 & 76.64  \\
    \bottomrule
    \end{tabular}}
    \caption{EuroSAT}
    \end{subtable}
    \hfill
    \begin{subtable}[t]{.3\textwidth}
    \centering
    \resizebox{\linewidth}{!}{\begin{tabular}{lccc}
    \toprule
    & Base & New & H \\
    \midrule
    CLIP & 70.53 & 77.50 & 73.85 \\
    CoOp & 84.69 & 56.05 & 67.46 \\
    CoCoOp & 82.33 & 73.45 & 77.64 \\
    CLIPood & \sbest{85.70} & \textbf{79.30} & \sbest{82.40}    \\
    MaPLe & 83.00 & \sbest{78.66} & 80.77 \\
    \FT{} &  \textbf{87.44} & 77.99 & \textbf{82.45} \\
    \midrule
    \method{} & 85.21 & 78.37 & 81.65 \\
    \bottomrule
    \end{tabular}}
    \caption{UCF101}
    \end{subtable}
    \hfill
    \caption{Classification accuracy (\%) from base to new classes over 11 datasets. The \textbf{best} and \sbest{second best} results are marked.}
    \label{tab:on_new_classes_benchmark}
\end{table}


\subsection{Visualization of Image Retrieval}
\begin{figure*}[ht]

    \centering
    \includegraphics[width=0.49\textwidth]{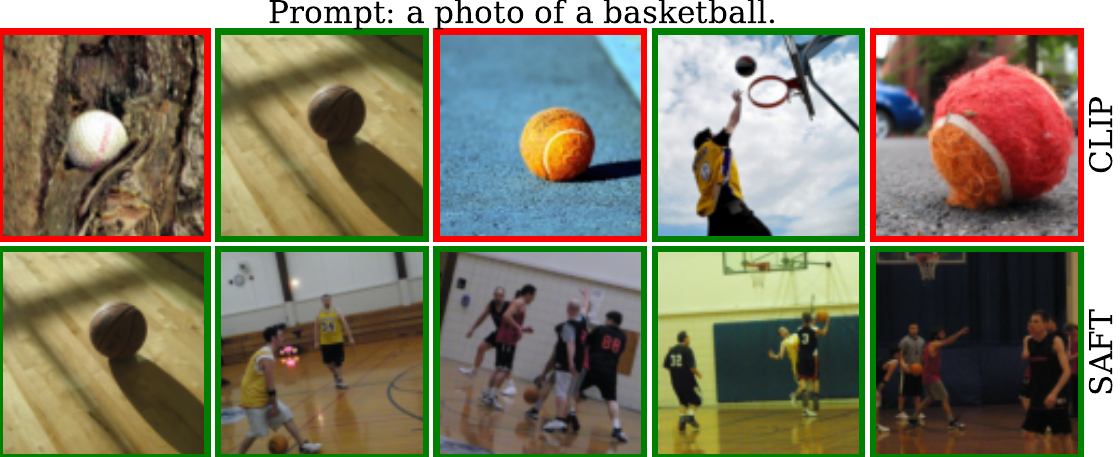}
    \includegraphics[width=0.49\textwidth]{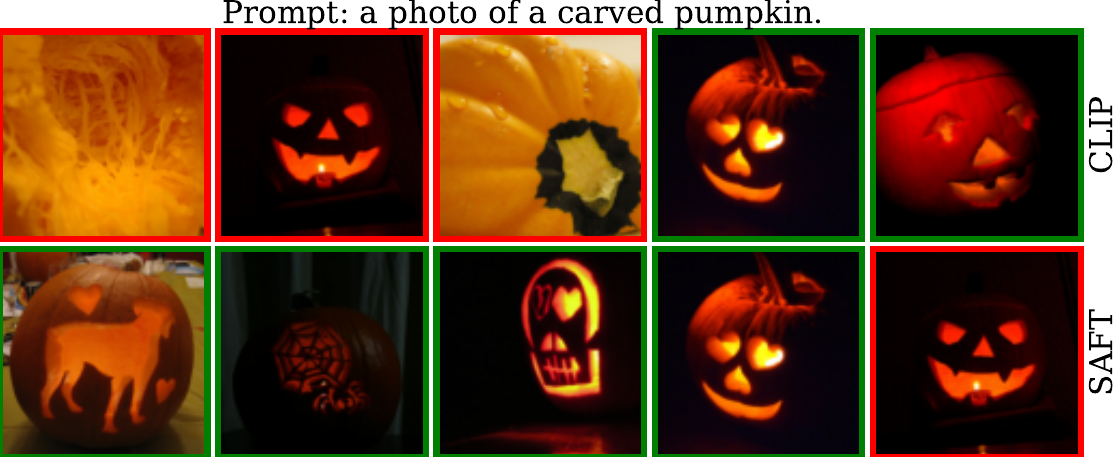}
    \vspace{5pt}
    
    \includegraphics[width=0.49\textwidth]{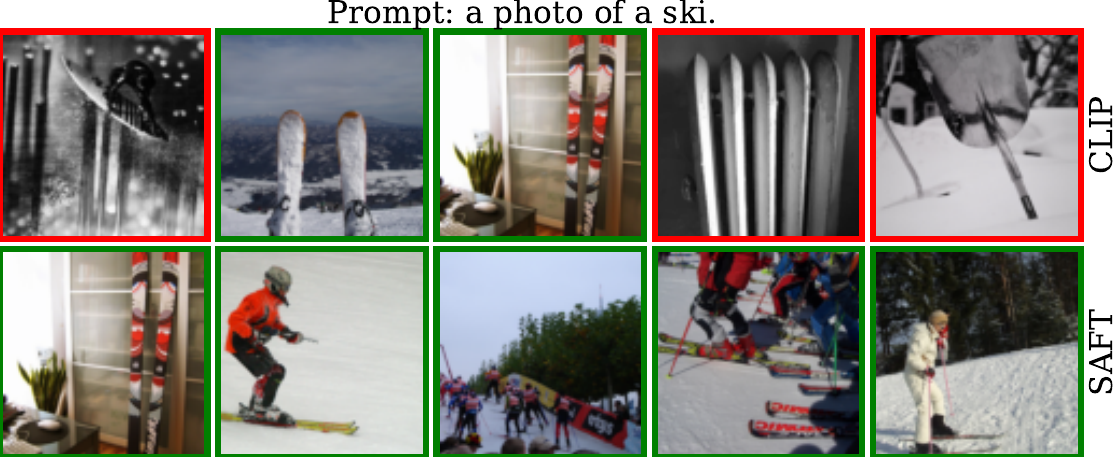}
    \includegraphics[width=0.49\textwidth]{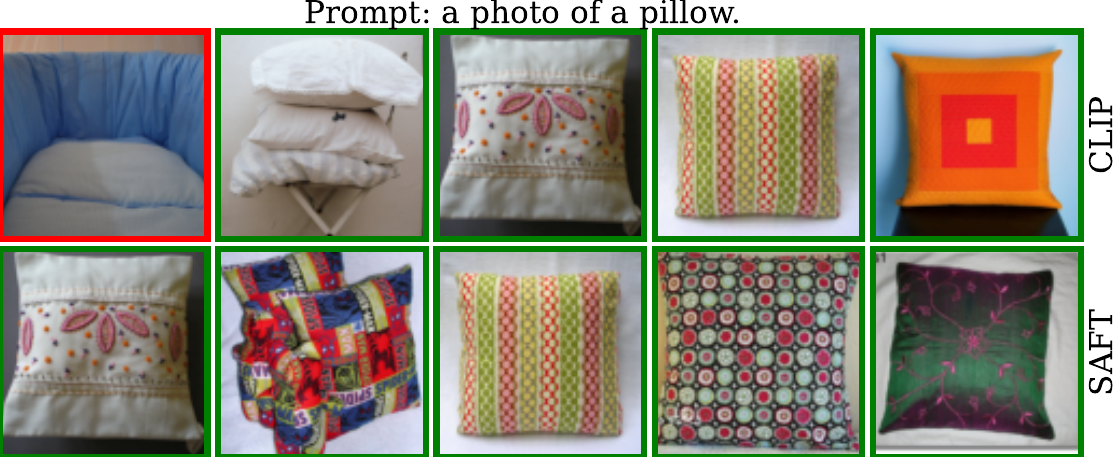}
    \vspace{5pt}

    \includegraphics[width=0.49\textwidth]{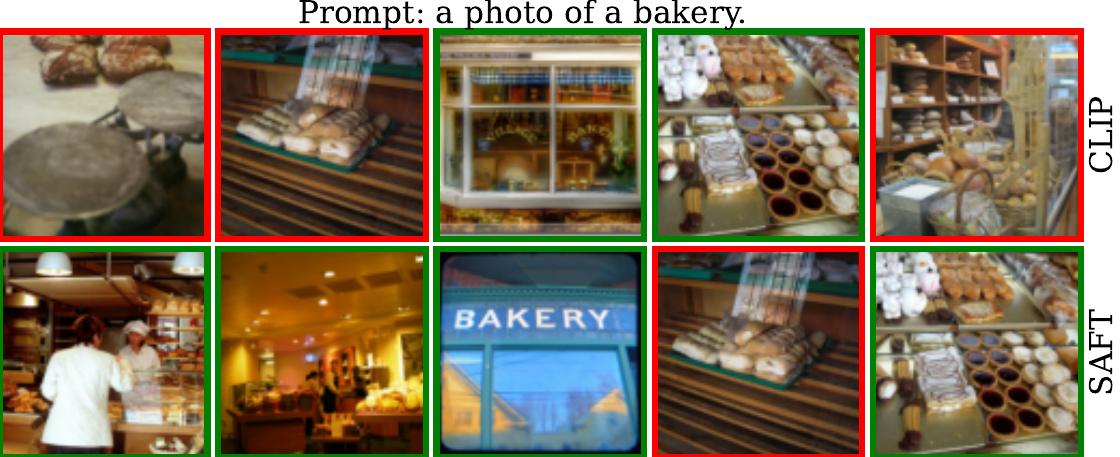}
    \includegraphics[width=0.49\textwidth]{figure/retrieval/hot_dog.pdf}
    \vspace{5pt}

    \includegraphics[width=0.49\textwidth]{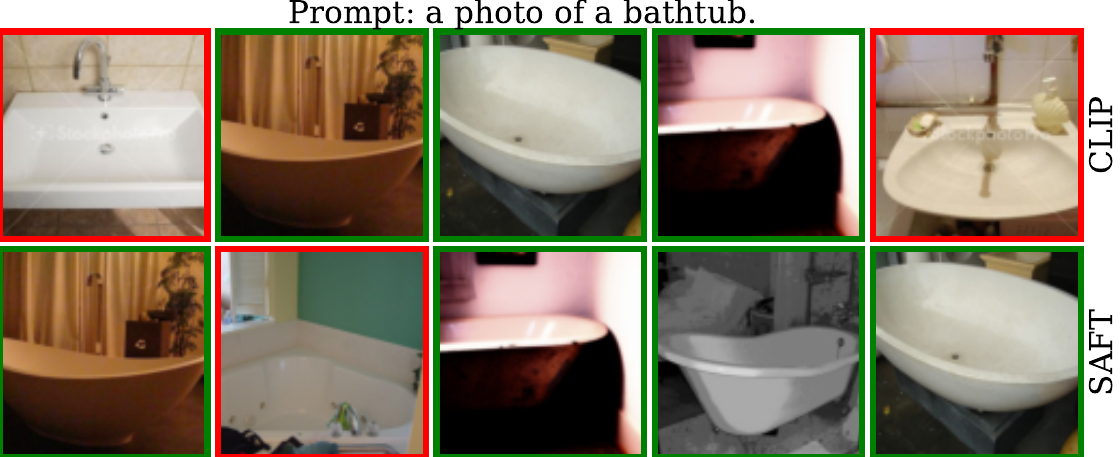}
    \includegraphics[width=0.49\textwidth]{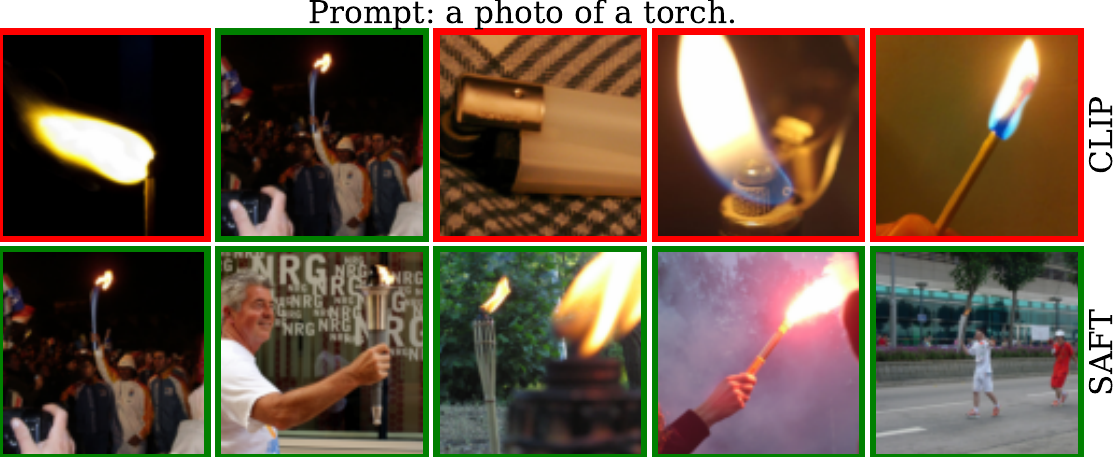}
    \vspace{5pt}

    \includegraphics[width=0.49\textwidth]{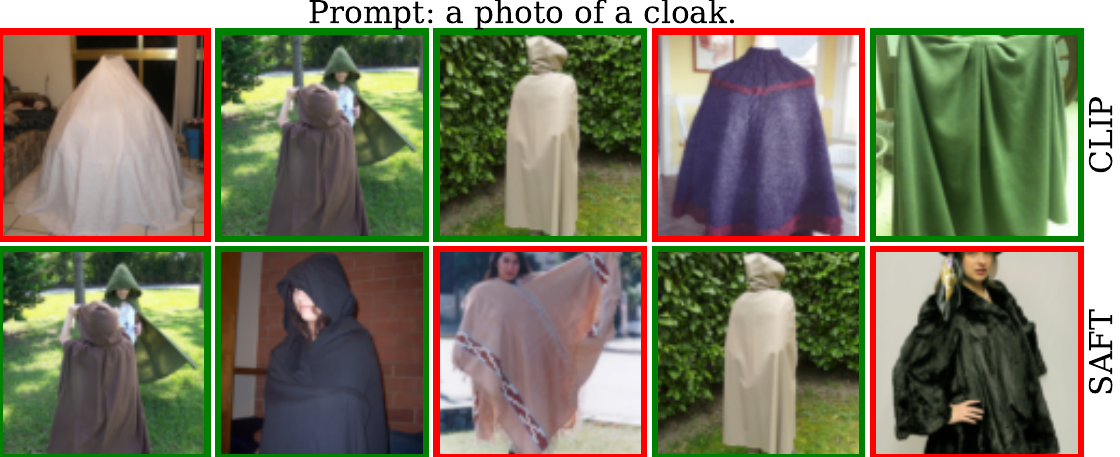}
    \includegraphics[width=0.49\textwidth]{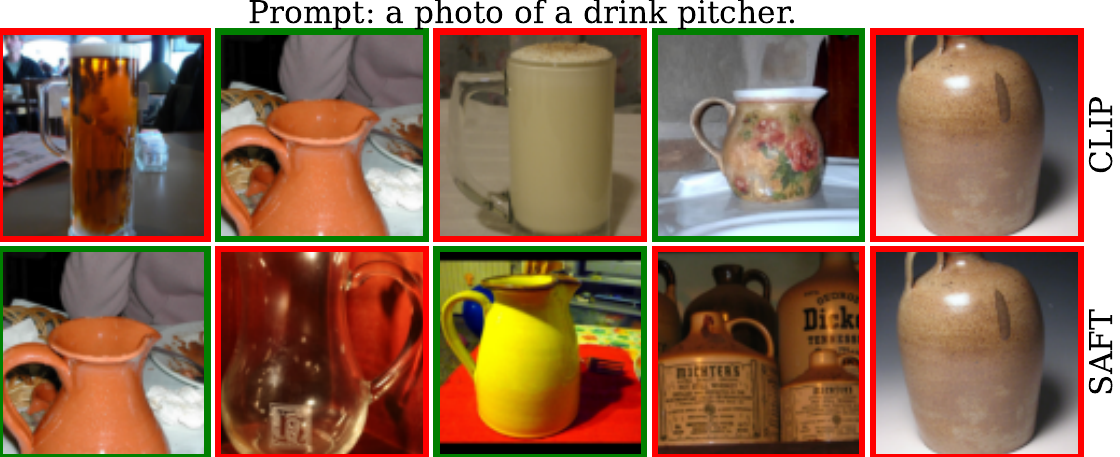}
    \vspace{5pt}
    
    \includegraphics[width=0.49\textwidth]{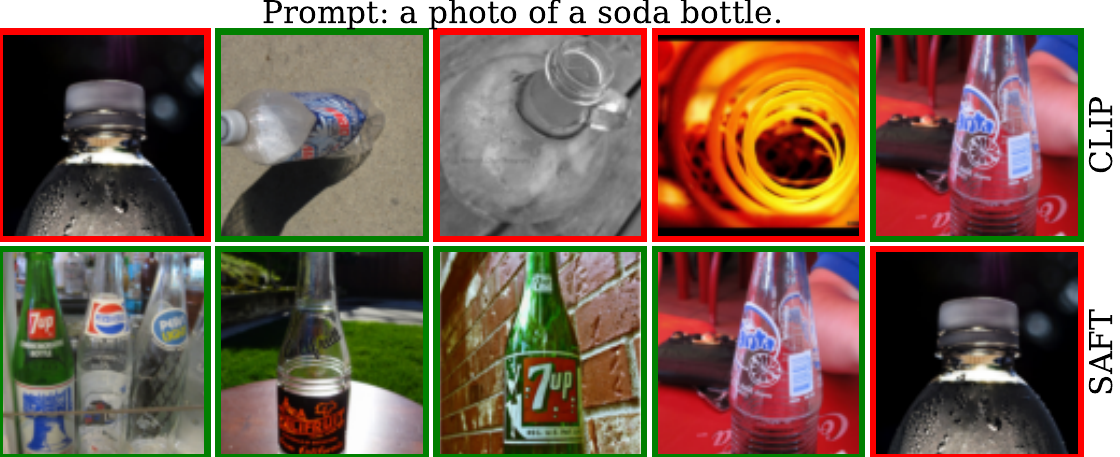}
    \includegraphics[width=0.49\textwidth]{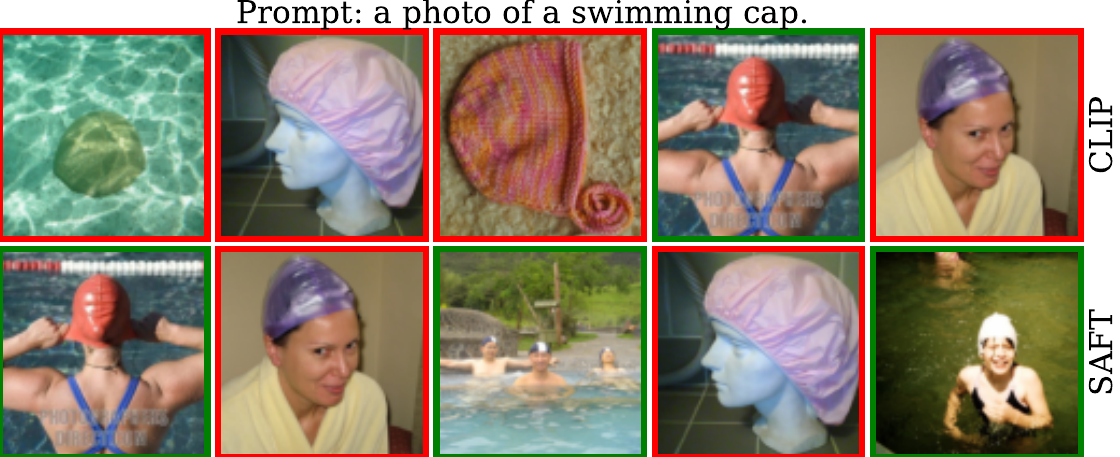}

    \caption{Top-5 retrieved images for a given prompt. Images are arranged from left to right in descending order of similarity to the given prompt. A green box indicates a correct match between the image and text, while a red box indicates an incorrect match.}
    \label{fig:top_5_retrieval}
    
\end{figure*}

In~\cref{fig:top_5_retrieval}, we show the top-5 image retrieval for given prompts. Using 16 shots per class from ImageNet for fine-tuning, \method{} outperforms CLIP by accurately retrieving similar images for a given prompt. The results demonstrate that fine-tuning can improve the performance of CLIP.

\subsection{Experiments on NLP Tasks} \label{appendix:nlp}
In this section, we further validate the OOD robustness of \method{} on various NLP tasks. Experiments are carried out using pre-trained large language models. Following Yuan \etal~\cite{yuan2023revisiting}, we use the BOSS\footnote{Available at: \url{https://github.com/lifan-yuan/OOD_NLP}} benchmark consisting of five tasks and twenty datasets. This benchmark contains a variety of NLP tasks, including sentiment analysis (SA), toxic detection (TD), and natural language inference (NLI) for classification; name entity recognition  (NER) for structured prediction; and extractive question answering (EQA) for reading comprehension. The text sources include product reviews, movie reviews, Twitter, and adversarial texts. On each task, one dataset is used for fine-tuning and ID evaluation, while three other datasets are used for OOD evaluation (see \cref{tab:boss}).
We consider T5-large~\cite{raffel2020exploring} and DeBERTa-large~\cite{he2020deberta} as backbone networks.  For NER, we add a classification head to DeBERTA which is trained on the ID data. This model is then used as an initialization for LoRA and SAFT. For the other tasks, we use manual pre-defined templates for fine-tuning. Our method \method{} is compared against the zero-shot pre-trained model, the conventional FT, and LoRA~\cite{hu2021lora}. We provide a detailed comparison of \method{} and LoRA with various sparsity levels in \cref{tab:boss:sa:full}.

In addition, we report the experimental results for fine-tuning a larger T5 model with 3B parameters. The results in \cref{tab:boss:sa:large} again confirm the effectiveness of \method{} for large-scale models.

\begin{table*}[!th]
\centering
\resizebox{0.8\linewidth}{!}{
\begin{tabular}{lllll}
\toprule
Task & \multicolumn{1}{c}{ID} & \multicolumn{3}{c}{OOD}                   \\ 
\cmidrule(lr){1-1} \cmidrule(lr){2-2} \cmidrule(lr){3-5}
SA   & Amazon (AZ)                     & Dynasent (DS) & SemEval (SE)       & SST (SST)     \\
TD   & Civil Comments (CC)             & AdvCivil (AC) & Implicit Hate (IH) & ToxiGen (TG)  \\
NLI  & MNLI (MN)                       & ANLI (AN)     & ContractNLI (CN)   & WANLI (WN)    \\
NER  & FewNerd (FN)                    & CoNLL (CoNLL) & E-NER (ENER)       & WNUT (WNUT)   \\
EQA  & SQuAD (SQuAD)                   & AdvQA (AQA)   & NewsQA (NQA)       & SearchQA (QA) \\ \bottomrule
\end{tabular}
}
\caption{Overview of datasets used in the BOSS benchmark}

\label{tab:boss}
\end{table*}

\subsection{Further Ablation Studies}
\textbf{Fine-tuning choice.} Another approach to mitigate overfitting during CLIP fine-tuning is to restrict the fine-tuned architecture. One option is to fine-tune the image encoder while keeping the text encoder frozen (\textit{Visual}), or conversely, fine-tuning the text encoder while keeping the image encoder frozen (\textit{Text}). \cref{tab:selection_strategy_1} shows that fine-tuning both encoders leads to better performance for our method. Fine-tuning solely the text encoder is more effective than fine-tuning the image encoder alone. Our findings are consistent with previous studies~\cite{zhai2022lit}.   

\begin{table}[!t]
\centering
\resizebox{0.8\linewidth}{!}{
\begin{tabular}{lccccc}
\toprule
 \multirow{3}*{Method} & ID & \multicolumn{4}{c}{OOD}\\
 \cmidrule(lr){2-2} \cmidrule(lr){3-6}
 & ImageNet & ImageNet-V2 & ImageNet-S & ImageNet-A & ImageNet-R \\
\midrule
\method{} & \textbf{72.71} & \textbf{65.84} & \textbf{49.69} & \textbf{51.60} & \textbf{78.13}  \\
\midrule
Visual & 70.61 & 63.97 & 48.84 & 48.96 & 77.56 \\
Text & 71.66 & 64.64 & 49.30 & 50.33 & 77.18  \\
\bottomrule
\end{tabular}}
\caption{Ablation studies on the ImageNet benchmark. The best results are in \textbf{bold}.}
\label{tab:selection_strategy_1}
\vspace{-10pt}
\end{table}

\textbf{Parameter selection strategy.} We further investigate the importance of the selection strategy presented in~\cref{eq:importance} to determine the parameters for updating. Another commonly used strategy, found in the literature, involves approximating the diagonal elements of the Fisher information matrix~\cite{kirkpatrick2017overcoming,sung2021training}. This strategy, denoted as SAFT$^\dagger$, uses
\begin{equation}
\vw = \frac{1}{N} \sum\nolimits_{i=1}^N \left(\nabla_{\theta} \mathcal{L}_{\text{CE}}(\vx_i, y_i; \theta)\right)^2
\end{equation}
to select the parameters that should be updated. In~\cref{table:openclass_aux}, we compare the performance of our method (\method{}) against SAFT$^\dagger$ in terms of generalization from base to new classes, as outlined in~\cref{sub:generalization_new_classes}. Additionally, in~\cref{tab:selection_strategy_aux}, we present the results of both \method{} and \method{}$^\dagger$ on the ImageNet benchmark, as described~\cref{sub:domain_shift}. Despite the fact that these methods only differ in their parameter selection strategy, our method most of the times outperforms SAFT$^\dagger$ across various scenarios. These findings underscore the effectiveness of our proposed approach.

\begin{table}[t]
\centering
\begin{tabular}{lcccccc}
\toprule
\multirow{3}{*}{Method}   & \multicolumn{3}{c}{\textit{Average}} & \multicolumn{3}{c}{ImageNet} \\
\cmidrule(lr){2-4} \cmidrule(lr){5-7}
 & Base  & New   & H   & Base  & New   & H     \\
\midrule
\method{} & 83.97 &  \textbf{74.78} &  \textbf{79.11} & \textbf{77.74}  & \textbf{71.40} & \textbf{74.44} \\
\method{}$^\dagger$ & \textbf{84.09} &  74.34 &  78.92 & 77.34  & 71.04 & 74.06 \\
\bottomrule                          
\end{tabular}
\caption{Classification accuracy (\%) from base to new classes over 11 datasets. The \textbf{best} results are marked.}
\label{table:openclass_aux}
\end{table}

\begin{table}[t]
\centering
\begin{tabular}{lccccc}
\toprule
 \multirow{3}*{Method} & ID & \multicolumn{4}{c}{OOD}\\
 \cmidrule(lr){2-2} \cmidrule(lr){3-6}
 & ImageNet & ImageNet-V2 & ImageNet-S & ImageNet-A & ImageNet-R \\
\midrule
\method{} & \textbf{72.71} & \textbf{65.84} & \textbf{49.69} & \textbf{51.60} & \textbf{78.13}  \\
\method{}$^\dagger$ & 72.22 & 65.32 & 49.50 & 51.07 & 77.92 \\
\bottomrule
\end{tabular}
\caption{Classification accuracy (\%) on the ImageNet benchmark. The \textbf{best} results are marked.}
\label{tab:selection_strategy_aux}
\vspace{-10pt}
\end{table}

\begin{table}[!hb]
\vspace{-10pt}
\tablestyle{9pt}{1.1}
\renewcommand{\arraystretch}{0.99}
\resizebox{\linewidth}{!}{
\begin{tabular}{lllrrccccc}
\toprule
\multirow{3}{*}{Task} &\multirow{3}{*}{Metric}   & \multirow{3}{*}{Method} & \multirow{3}{*}{\shortstack{\# fine-tuned\\ parameters}} & \multirow{3}{*}{Ratio} & \multicolumn{5}{c}{Datasets} \\
\cmidrule(lr){6-10}
                 &  &  &              &   & ID                    & \multicolumn{4}{c}{OOD} \\
\midrule
\multirow{5}{*}{NER} & \multirow{5}{*}{F1}  &       &                           &        & FN & CoNLL & ENER & WNUT & \textit{Average} \\
\cmidrule(lr){6-6} \cmidrule(lr){7-10}
                     &   & DeBERTa (large) & 17.43K  & 0.004\% & 52.9 & 60.0 & 36.8 & 29.5 & 42.1 \\
                     &   & FT              & 434.03M & 100\%   & \textbf{79.4} & 70.6 & 50.0 & 43.2 & 54.6 \\ \cmidrule(lr){3-10}
                     &   & LoRA ($r = 1$)  & 133.15K & 0.031\% & 75.7 & 67.8 & 56.7 & 43.5 & 56.0 \\
                     &   & LoRA ($r = 4$)  & 428.07K & 0.099\% & 77.1 & 66.3 & 57.2 & 45.1 & 56.2 \\
                     &   & LoRA ($r = 8$)  & 821.28K & 0.189\% & 77.7 & 67.5 & 57.2 & 45.3 & \textbf{56.6} \\
                     \cmidrule(lr){3-10}
                     &   & SAFT ($\alpha=0.00025$) &  108.57K & 0.025\% & 76.6 & 67.4 & 55.7 & 46.3 & \sbest{56.5} \\
                     &   & SAFT ($\alpha=0.0005$)  &  217.07K & 0.050\% & 77.5 & 66.5 & 53.6 & 47.1 & 55.7 \\
                     &   & SAFT ($\alpha=0.001$)   &  434.08K & 0.100\% & \sbest{78.3} & 66.7 & 54.0 & 47.8 & 56.2 \\
\midrule
\multirow{5}{*}{SA} & \multirow{5}{*}{Accuracy}    &    &                           &        & AZ	& DS&	SE&	SST & \textit{Average} \\
\cmidrule(lr){6-6} \cmidrule(lr){7-10}
                      & & T5 (large)     & 0        & 0\%    & 85.78	& 35.58 & 34.55 & 43.49	& 37.87                  \\
                      & & FT             & 737.67M  & 100\%  & \textbf{91.22} & 46.69 & 46.68 & 75.07 & 56.15                  \\  \cmidrule(lr){3-10}
                      & & LoRA ($r = 1$) & 294.91K & 0.040\% & 88.65 & 48.33 & 49.31 & 74.13 & 57.26 \\
                      & & LoRA ($r = 4$) & 1,179.65K & 0.160\% & 89.67 & 47.66 & 48.96 & 76.38 & 57.67 \\
                      & & LoRA ($r = 8$) & 2,359.30K & 0.319\% & \sbest{90.37} & 47.20 & 48.81 & 77.13 & 57.71 \\
                      \cmidrule(lr){3-10}
                      & & SAFT ($\alpha=0.00025$) & 184.42K & 0.025\% & 90.06 & 48.61 & 50.30 & 75.35 & 58.09 \\
                      & & SAFT ($\alpha=0.0005$)  & 368.83K & 0.050\% & 89.62 & 48.75 & 49.66 & 76.48 & \textbf{58.29} \\
                      & & SAFT ($\alpha=0.001$)   & 737.67K & 0.100\% & 89.78 & 48.68 & 50.13 & 76.01 & \sbest{58.27} \\
\midrule
\multirow{5}{*}{TD} &  \multirow{5}{*}{Accuracy} &      &                           &        & CC	& AC	& IH &	TG & \textit{Average} \\
\cmidrule(lr){6-6} \cmidrule(lr){7-10}
                      &  & T5 (large)    & 0       & 0\%     & 14.90 & 79.47 & 39.54 & 43.94 & 54.31 \\
                      &  & FT            & 737.67M & 100\%   & 85.43 & 64.28 & 62.47 & 68.94 & 65.23 \\  \cmidrule(lr){3-10}
                      & & LoRA ($r = 1$) & 294.91K & 0.040\% & 84.31 & 59.54 & 60.22 & 64.89 & 61.55 \\
                      & & LoRA ($r = 4$) & 1,179.65K & 0.160\% & \sbest{86.17} & 68.29 & 60.80 & 65.21 & 64.77 \\
                      & & LoRA ($r = 8$) & 2,359.30K & 0.319\% & \textbf{86.22} & 69.14 & 60.55 & 64.15 & 64.61 \\
                      \cmidrule(lr){3-10}
                      & & SAFT ($\alpha=0.00025$) & 184.42K & 0.025\% & 85.63 & 71.81 & 62.17 & 64.36 & \sbest{66.11} \\
                      & & SAFT ($\alpha=0.0005$)  & 368.83K & 0.050\% & 85.31 & 73.15 & 61.94 & 64.89 & \textbf{66.66} \\
                      & & SAFT ($\alpha=0.001$)   & 737.67K & 0.100\% & 84.72 & 68.77 & 60.76 & 65.21 & 64.92 \\
\midrule
\multirow{5}{*}{NLI} &   \multirow{5}{*}{Accuracy}&      &                           &        & MN &	AN	& CN &	WN & \textit{Average} \\
\cmidrule(lr){6-6} \cmidrule(lr){7-10}
                      &  & T5 (large)    & 0       & 0\%     & 35.02 & 33.03 &	45.72 &	19.22 &	32.66                  \\
                      &  & FT            & 737.67M & 100\%   & \textbf{89.25} & 37.19	& 38.79	& 62.66	&	46.21                  \\  \cmidrule(lr){3-10}
                      & & LoRA ($r = 1$) & 294.91K & 0.040\% & 87.62 & 29.78 & 42.66 & 57.72 & 43.39 \\
                      & & LoRA ($r = 4$) & 1,179.65K & 0.160\% & 88.85 & 32.63 & 47.35 & 59.50 & 46.49 \\
                      & & LoRA ($r = 8$) & 2,359.30K & 0.319\% & \sbest{89.21} & 33.44 & 46.63 & 60.26 & 46.77 \\
                      \cmidrule(lr){3-10}
                      & & SAFT ($\alpha=0.00025$) & 184.42K & 0.025\% & 86.65 & 30.28 & 57.44 & 56.82 & \textbf{48.18} \\
                      & & SAFT ($\alpha=0.0005$)  & 368.83K & 0.050\% & 87.60 & 30.69 & 53.56 & 57.92 & \sbest{47.39} \\
                      & & SAFT ($\alpha=0.001$)   & 737.67K & 0.100\% & 88.13 & 31.34 & 50.65 & 59.06 & 47.02 \\
\midrule
\multirow{5}{*}{EQA} &  \multirow{5}{*}{F1}&      &                           &        & SQuAD	& AQA	& NQA	& SQA & \textit{Average} \\
\cmidrule(lr){6-6} \cmidrule(lr){7-10}
                      &  & T5 (large)    & 0       & 0\%     & 27.41 &	10.14 &	29.96	& 21.23		& 20.45             \\
                      & & FT             & 737.67M & 100\%   & \sbest{93.36} &	49.93 &	64.36	& 38.27	&	50.85    \\  \cmidrule(lr){3-10}
                      & & LoRA ($r = 1$) & 294.91K & 0.040\% & 93.22 & 49.99 & 65.94 & 37.65 & 51.20 \\
                      & & LoRA ($r = 4$) & 1,179.65K & 0.160\% & \textbf{93.38} & 50.88 & 66.06 & 38.99 & 51.98 \\
                      & & LoRA ($r = 8$) & 2,359.30K & 0.319\% & 93.33 & 50.52 & 65.42 & 36.89 & 50.94 \\
                      \cmidrule(lr){3-10}
                      & & SAFT ($\alpha=0.00025$) & 184.42K & 0.025\% & 93.11 & 50.34 & 66.61 & 40.26 & \textbf{52.40} \\
                      & & SAFT ($\alpha=0.0005$)  & 368.83K & 0.050\% & 93.11 & 50.18 & 66.40 & 40.04 & \sbest{52.21} \\
                      & & SAFT ($\alpha=0.001$)   & 737.67K & 0.100\% & 93.19 & 50.26 & 65.86 & 38.42 & 51.51 \\
                    \bottomrule
\end{tabular}
}
\caption{Results on the BOSS benchmark. The \textbf{best} and \sbest{second best} ID and OOD averages are marked.} 
\label{tab:boss:sa:full}
\end{table}

\begin{table}[!hb]
\vspace{-10pt}
\tablestyle{9pt}{1.1}
\renewcommand{\arraystretch}{0.99}
\resizebox{\linewidth}{!}{
\begin{tabular}{lllrrccccc}
\toprule
\multirow{3}{*}{Task} &\multirow{3}{*}{Metric}   & \multirow{3}{*}{Method} & \multirow{3}{*}{\shortstack{\# fine-tuned\\ parameters}} & \multirow{3}{*}{Ratio} & \multicolumn{5}{c}{Datasets} \\
\cmidrule(lr){6-10}
                 &  &  &              &   & ID                    & \multicolumn{4}{c}{OOD} \\
\midrule
\multirow{5}{*}{SA} & \multirow{5}{*}{Accuracy}    &    &                           &        & AZ	& DS&	SE&	SST & \textit{Average} \\
\cmidrule(lr){6-6} \cmidrule(lr){7-10}
                      & & T5 (3B)     & 0        & 0\%   & 84.53 &	33.63 &	34.27 &	37.68	&	35.20                  \\
                      & & FT   & 2,851.60M  & 100\%  & 91.35 & 50.74 & 50.89	& 75.07 & 58.90                  \\  \cmidrule(lr){3-10}
                      & & LoRA ($r = 1$) & 737.28K &	0.026\%	& 90.59 &	50.09 &	48.27 &	\sbest{77.79} &	58.72 \\
                      & & LoRA ($r = 4$) &  2.95M &	0.103\%	& 90.66 &	\textbf{52.04} &	50.60 &	77.32 &	59.99 \\
                      & & LoRA ($r = 8$) & 5.90M &	0.206\% & \textbf{92.02} &	51.83 &	\sbest{51.09} &	\textbf{77.98} &	\sbest{60.30} \\
                      \cmidrule(lr){3-10}
                      & & SAFT ($\alpha=0.00005$) & 142.58K	& 0.005\%& 91.05 &	49.82 &	49.56 &	76.85 & 58.74 \\
                      & & SAFT ($\alpha=0.0001$)  &  285.16K &	0.010\% &	91.04 &	50.51 &	49.93 &	77.69 &	59.38\\
                      & & SAFT ($\alpha=0.001$)   &  2.85M &	0.100\%	& \sbest{91.89} &	\sbest{51.94} &	\textbf{52.89} &	77.69 &	\textbf{60.84}\\                     
\midrule
\multirow{5}{*}{TD} &  \multirow{5}{*}{Accuracy} &      &                           &        & CC	& AC	& IH &	TG & \textit{Average} \\
\cmidrule(lr){6-6} \cmidrule(lr){7-10}
                      &  & T5 (3B)    & 0       & 0\%     & 21.04 &	75.70 &	40.72 &	44.04 &	42.38\\
                      &  & FT            & 2,851.60M  & 100\% & \textbf{87.22} & 64.52 & \textbf{63.51} & \textbf{70.85} &	67.18 \\  \cmidrule(lr){3-10}
                      & & LoRA ($r = 1$) & 737.28K &	0.026\%	& 83.61 & \textbf{78.74} & 60.25	& \sbest{67.98}	&	\textbf{68.99} \\
                      & & LoRA ($r = 4$) &  2.95M &	0.103\%	& 85.39 & 75.33 & 61.12 & 67.13 & \sbest{67.86}\\
                      & & LoRA ($r = 8$) & 5.90M &	0.206\% &  \sbest{86.36} &	75.09 & \sbest{61.96} &	66.70 &	67.92 \\
                      \cmidrule(lr){3-10}
                       & & SAFT ($\alpha=0.00005$) & 142.58K	& 0.005\%& 83.32 & \sbest{76.67} & 60.62 & 65.85 & 67.71 \\
                      & & SAFT ($\alpha=0.0001$)  &  285.16K &	0.010\% & 82.92 & \sbest{76.67} & 59.72 & 66.60 &67.66	\\
                      & & SAFT ($\alpha=0.001$)   &  2.85M &	0.100\%	& 86.13 & 70.72 & 60.95 & 66.38 & 66.02 \\ 
\midrule
\multirow{5}{*}{EQA} &  \multirow{5}{*}{F1}&      &                           &        & SQuAD	& AQA	& NQA	& SQA & \textit{Average} \\
\cmidrule(lr){6-6} \cmidrule(lr){7-10}
                      &  & T5 (3B)    & 0       & 0\%     & 46.64	&18.88 & 21.36	& 13.42 & 17.89 \\
                      & & FT             & 2,851.60M  & 100\%   &  94.33 &	57.01 &	65.44 &	26.60 &	49.68\\  \cmidrule(lr){3-10}
                      & & LoRA ($r = 1$) & 737.28K &	0.026\%	& \textbf{94.77} & 58.56 & \textbf{67.16} &	21.88 & 49.20 \\
                      & & LoRA ($r = 4$) &  2.95M &	0.103\%	& 94.70 & 58.92 & 66.03 &	21.97 & 48.98 \\
                      & & LoRA ($r = 8$) & 5.90M &	0.206\% & \sbest{94.71} & \sbest{59.21} & 65.68	& 22.88 & 49.26 \\
                      \cmidrule(lr){3-10}
                      & & SAFT ($\alpha=0.00005$) & 142.58K	& 0.005\%& 94.36 &	57.68 &	\sbest{66.82} & \textbf{29.77}	& \textbf{51.42}  \\
                      & & SAFT ($\alpha=0.0001$)  &  285.16K &	0.010\% & 94.43 &	58.14 & 66.74 & \sbest{26.85} & \sbest{50.58}	\\
                      & & SAFT ($\alpha=0.001$)   &  2.85M &	0.100\%	& 94.61 & \textbf{59.38} &	65.47 & 20.57 & 48.47\\
                    \bottomrule
\end{tabular}
}
\caption{Results on the BOSS benchmark. The \textbf{best} and \sbest{second best} ID and OOD averages are marked.} 
\label{tab:boss:sa:large}
\end{table}

\end{document}